\documentclass[acmsmall]{acmart}

\usepackage{graphicx}
\usepackage{amsmath}
\usepackage{multirow}
\usepackage{caption} 
\usepackage{subcaption}
\usepackage{xcolor}

\usepackage{algorithm}
\usepackage[noend]{algorithmic}

\renewcommand{\algorithmiccomment}[1]{\bgroup\hfill//~#1\egroup}
\usepackage{longtable}
\usepackage{array}
\usepackage{booktabs}
\usepackage{caption}
\usepackage{color, colortbl}
\usepackage{multirow}
\usepackage{diagbox}
\usepackage{algorithm}
\usepackage{listings}
\definecolor{rc1}{RGB}{235,235,235}
\definecolor{rc2}{RGB}{255,255,255}
\definecolor{codeblue}{rgb}{0.25,0.5,0.5}
\definecolor{codekw}{rgb}{0.85, 0.18, 0.50}

\newcommand{\specialcellleft}[2][l]{%
\begin{tabular}[#1]{@{}l@{}}#2\end{tabular}}

\usepackage{fontawesome}

\usepackage{microtype}

\usepackage{etoolbox}
\makeatletter
\AfterEndEnvironment{algorithm}{\let\@algcomment\relax}
\AtEndEnvironment{algorithm}{\kern2pt\hrule\relax\vskip3pt\@algcomment}
\let\@algcomment\relax
\newcommand\algcomment[1]{\def\@algcomment{\footnotesize#1}}
\renewcommand\fs@ruled{\def\@fs@cfont{\bfseries}\let\@fs@capt\floatc@ruled
  \def\@fs@pre{\hrule height.8pt depth0pt \kern2pt}%
  \def\@fs@post{}%
  \def\@fs@mid{\kern2pt\hrule\kern2pt}%
  \let\@fs@iftopcapt\iftrue}
\makeatother

\usepackage{pifont}
\newcommand{\cmark}{\checkmark} 
\newcommand{\xmark}{\ding{55}} 
\usepackage{graphicx}
\usepackage{amsmath}
\usepackage{booktabs}
\usepackage{enumitem}

\AtBeginDocument{%
  }

\setcopyright{acmcopyright}
\acmDOI{XXXXXXX.XXXXXXX}

\acmJournal{TOMM}

\begin{document}

\newcommand{\blocka}[2]{\multirow{3}{*}{\(\left[\begin{array}{c}\text{3$\times$3, #1}\\[-.1em] \text{3$\times$3, #1} \end{array}\right]\)$\times$#2}
}

\newcommand{\blockb}[3]{\multirow{3}{*}{\(\left[\begin{array}{c}\text{1$\times$1, #2}\\[-.3em] \text{3$\times$3, #2}\\[-.3em] \text{1$\times$1, #1}\end{array}\right]\)$\times$#3}
}


\title{Contrastive Learning of View-Invariant Representations for Facial Expressions Recognition}

\author{Shuvendu Roy}
\email{shuvendu.roy@queensu.ca}
\orcid{0002-7674-0238}
\author{Ali Etemad}
\email{ali.etemad@queensu.ca}
\affiliation{%
  \institution{\\Dept. ECE and Ingenuity Labs Research Institute, 
Queen's University}
  \city{Kingston}
  \state{Ontario}
  \country{Canada}
}

\renewcommand{\shortauthors}{Roy et al.}
 
\begin{abstract}
Although there has been much progress in the area of facial expression recognition (FER), most existing methods suffer when presented with images that have been captured from viewing angles that are non-frontal and substantially different from those used in the training process. In this paper, we propose ViewFX, a novel view-invariant FER framework based on contrastive learning, capable of accurately classifying facial expressions regardless of the input viewing angles during inference. ViewFX learns view-invariant features of expression using a proposed self-supervised contrastive loss which brings together different views of the same subject with a particular expression in the embedding space. We also introduce a supervised contrastive loss to push the learnt view-invariant features of each expression away from other expressions. Since facial expressions are often distinguished with very subtle differences in the learned feature space, we incorporate the Barlow twins loss to reduce the redundancy and correlations of the representations in the learned representations. The proposed method is a substantial extension of our previously proposed CL-MEx, which only had a self-supervised loss. We test the proposed framework on two public multi-view facial expression recognition datasets, KDEF and DDCF. The experiments demonstrate that our approach outperforms previous works in the area and sets a new state-of-the-art for both datasets while showing considerably less sensitivity to challenging angles and the number of output labels used for training. We also perform detailed sensitivity and ablation experiments to evaluate the impact of different components of our model as well as its sensitivity to different parameters.
\end{abstract}

\ccsdesc[500]{Computing methodologies~Image representations}
\ccsdesc[500]{Computing methodologies~Unsupervised learning}

\keywords{Affective computing, Contrastive Learning, Expression Recognition}


\maketitle

\section{Introduction}\label{sec:introduction}

Facial expressions are a crucial form of non-verbal communication. As a result, facial expression recognition (FER) can play a key role in human-machine interaction systems by allowing the system to understand and adapt to human reactions. Examples of such systems include health-care assistants \cite{tokuno2011usage}, driving assistants \cite{leng2007experimental}, personal mood management systems \cite{thrasher2011mood, sanchez2013inferring}, emotion-aware multimedia \cite{cho2019instant}, and others. Nonetheless, FER remains a challenging task due to a number of issues, such as the subtlety of facial actions that result in the display of expressions \cite{samal1992automatic}. For instance, it has been well documented that facial expressions are determined by very \textit{subtle} differences in the face caused by intricate muscles \cite{samal1992automatic}. This results in higher than normal correlations between different classes (expressions) in the learned feature space. Hence, solutions capable of learning compact and de-correlated representations are required to achieve robust FER.

In practical settings, it is highly likely for images of human faces to be captured from different angles and not always adhere to a clear and standard frontal view. The appearance of facial expressions can vary in different viewpoints, posing unique challenges especially due to discrepancies between the extreme angles, e.g., profile versus frontal views. To leverage the plurality of facial views for FER, a class of solutions called multi-view FER has been evolved and shown to generate more robust results \cite{eleftheriadis2014discriminative, SVM, zhang2020geometry}. This itself leads to a new challenging category named view-\textit{invariant} FER, which aims to leverage multiple views during training but generate predictions at inference based solely on a single view \cite{eleftheriadis2014discriminative,wang2016facial,taheri2011towards}. In essence, the goal here is to be able to classify expressions irrespective of the input viewing angle at inference. 

\begin{figure}[t]
\centering
\includegraphics[width=0.6\columnwidth]{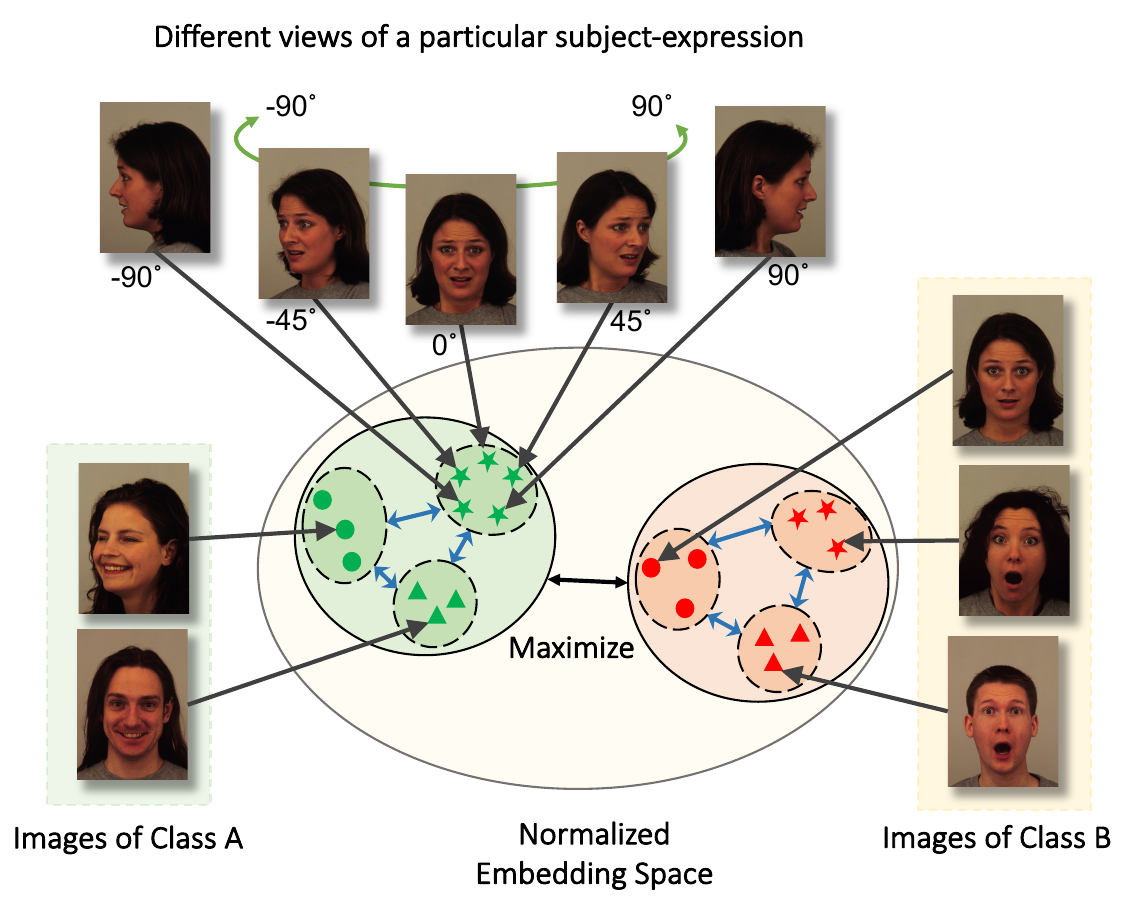} 
\caption{A visual overview of our proposed method, ViewFX. Different views of the same image are brought close to form view-based clusters, while external clusters are formed for each expression class.}
\label{fig:embedding}
\end{figure}

In this paper, to effectively perform view-invariant FER, we propose a novel solution based on contrastive learning \cite{simclr,supcon}, which we name ViewFX. Our method learns view-invariant expression representations by using a self-supervised contrastive approach to bring together different views of the same subject exhibiting a particular expression. It then uses a supervised contrastive loss to push the learned clusters away from other clusters in the embedding space that contain all the views of other expression classes. A high-level view of our method is depicted in Figure \ref{fig:embedding}. Further, in order to reduce the redundancy and correlations in the learned representation space and generate more compact embeddings, we incorporate the Barlow twins \cite{barlowTwin} as an additional loss term in our model. We rigorously evaluate our proposed method on two public multi-view FER datasets. Our experiments show that ViewFX achieves state-of-the-art results on both datasets, outperforming previous works in this area. Our model also shows a lower deviation in performance when tested against very challenging views in comparison to a similar encoder trained without the proposed method. We also carefully investigate the influence of different parameters and augmentations used in our pipeline. Finally, our ablation experiments demonstrate the positive impact of each of the different components of our model.
The proposed ViewFX is an extension of our CL-MEx that only employed a self-supervised contrastive loss.
Our main contributions are summarized as follows:
\begin{itemize}
    \item We propose a new framework for FER that learns effective and robust \textit{view-invariant} representations of facial expressions. 
    
    \item Our model uses a novel loss consisting of three terms, each of which improves a particular aspect of the model for downstream expression recognition. First, a self-supervised contrastive term aims to learn view-invariant representations, followed by a supervised contrastive term aiming to discriminate between expression classes. Finally, the third term, the Barlow twins, aims to reduce ambiguities in the learned representations;

    \item We perform detailed experiments on two publicly available multi-view datasets, for which our model sets new state-of-the-art. Ablation studies validate the impact of each of the proposed loss terms. Additional experiments reveal the robustness of our method towards extreme viewing angles.
    
\end{itemize}

The rest of this paper is organized as follows. In the following section, we review the related literature on FER on both single-view and multi-view settings, as well as general contrastive self-supervised learning. This is followed by a detailed description of the problem setup, proposed method, network architecture, and training details. Next, the experiment setup and results are discussed. Lastly, in the final section, we present the concluding remarks, limitations, and future works.

\section{Related Works}\label{sec:related_work}
In this section, we review the related works in two key areas related to the proposed method: (1) FER solutions and (2) contrastive learning.  

\subsection{Facial Expression Recognition}
We discuss the previous works in three different categories: Fully supervised FER methods, Non-supervised methods, and View-invariant methods. 

\noindent \textbf{Supervised methods.} Recent works have focused on FER with deep learning by means of specifically designed architectures \cite{vo2020pyramid, hasani2020breg,tang2020facial}, attention modules \cite{wang2020region}, loss functions \cite{vo2020pyramid}, and training setups \cite{siqueira2020efficient, pourmirzaei2021using}, showing great improvements in performance versus classical machine learning solutions \cite{shan2005robust,essa1995facial,shan2009facial}. An example of a specially designed deep neural architecture for FER is BReG-NeXt \cite{hasani2020breg}, which introduced a modified residual connection \cite{resnet} in the network. Their proposed method replaces the standard residual connection with adaptive complex mapping, which helps boost the performance of the model while maintaining a shallower architecture with fewer computational requirements. An ensemble of models was proposed for FER in \cite{siqueira2020efficient}. It proposed a shared representation-based CNN to increase the efficiency of training the ensemble. Multi-modal models have also been explored in works such as \cite{schoneveld2021leveraging}, which have resulted in strong performances for expression recognition. 

Modifications to loss functions have also been shown to aid and improve the performance of deep FER architectures. For example, a prior distribution label smoothing loss was proposed in \cite{vo2020pyramid}, which added prior knowledge about the expressions into the classifier. In \cite{kolahdouzi2021face}, a new loss term was added to the FER pipeline to learn the optimum path of connecting extracted facial landmarks for creating face trees. The optimized face trees were subsequently used for learning both the texture and structure of the face toward FER, showing performance improvements versus prior works. Ad-Corre \cite{fard2022ad} proposed a metric learning approach to handle high intra-class variations and inter-class similarities of in-the-wild facial images. CFC-SP \cite{xue2022coarse} proposed a video facial expression recognition method that captures universal and unique expression features to boost recognition performance. In \cite{psaroudakis2022mixaugment}, a MixUp-based data augmentation module was proposed to reduce the data scarcity issue.

\noindent \textbf{Non-supervised methods.}
In addition to traditional fully supervised FER, there have been efforts towards self-supervised and semi-supervised learning based methods as well \cite{jiang2021boosting,ST-CLR,jiang2021boosting}. For instance, in \cite{li2021self}, Self-supervised Exclusive-Inclusive Interactive Learning (SEIIL) was proposed to learn multi-label compound expression features in a self-supervised manner. In \cite{pourmirzaei2021using}, a self-supervised auxiliary task was proposed to learn better representations from the data, which showed improved performance in downstream FER. 
In \cite{li2022crs}, contrastive self-supervised learning was utilized to learn robust expression features from the images. 
In \cite{ST-CLR}, to classify expressions in videos, a FER solution was proposed based on self-supervised learning. The model used contrastive learning with a novel temporal sampling technique to learn expression features in videos.
In \cite{jiang2021boosting}, a semi-supervised method called Progressive Teacher was proposed to utilize large-scale unlabelled expression images alone with a standard labelled dataset to boost the performance. A semi-supervised method using mean-teacher was proposed in \cite{wang2021multi} that uses the teacher model to predict pseudo-labels for unlabelled data. The pseudo-labels are then used to train the student model, which results in a more robust FER model. Meta-Face2Exp \cite{zeng2022face2exp} proposed to utilize unlabelled data to tackle the class imbalance issue of FER datasets. Ada-CM \cite{li2022towards} proposed a semi-supervised method that effectively utilizes all samples in the unlabelled set.

\noindent \textbf{Multi-view and view-invariant methods.} 
To leverage the information available in various views captured simultaneously with several cameras, several multi-view FER solutions have been proposed. 
For instance, \cite{vo20193d} proposed a multi-view CNN architecture that takes three facial views as input and fuses their outputs to predict the expression label.
In \cite{hu2008multi}, a two-step solution was proposed, wherein the first step, the view angle of an input image was predicted, and then a view-specific model was used to predict the output label. A face frontalization technique was proposed in \cite{zhang2018spatially} where a frontal representation of an input facial image was generated with a generative model, and then the expression label was predicted on the frontal image.
Light-field images capture a scene from multiple directions, creating a form of multi-view sequence. In the context of FER, Light-field images have been used in the past. For example, hybrid CNN-LSTM architectures were proposed for FER in \cite{sepas2020facial,sepas2019deep}. In \cite{sepas2021capsfield}, capsule networks were used for FER, showing improvements over several baselines. Subsequently, in \cite{sepas2021multi, sepas2020long}, three LSTM variants were proposed to jointly learn different views for several downstream applications, including multi-view FER, resulting in further improvements when used with light-field images.

View-invariant FER is a sub-field of multi-view methods, where the goal is to learn \textit{view-invariant} features of expression from multi-view information but perform the inference step only from a single-view input. In \cite{MPCNN}, for instance, view-invariant feature learning was achieved for FER with a multi-channel pose-aware CNN (MPCNN). This method proposed three main components - a multi-channel feature extractor, joint multi-scale feature fusion, and a pose-aware feature detector. 
Later in \cite{PhaNet}, to recognize the expressions and jointly model the views at the same time, a hierarchical pose-adaptive attention network was proposed. The attention module could pay attention to view-invariant features at any angle, resulting in a more robust FER model. 
In \cite{wang2020region}, a customized CNN network named Region Attention Network (RAN) was proposed, where an attention module was incorporated to learn robust features from the data, learning pose- and occlusion-aware features. In ST-SE \cite{liu2023soft}, a new block was introduced to the model that can extract salient features of channels for pose-invariant FER.
Finally, a contrastive loss was proposed in CL-MEx \cite{CL_MEx} that could learn the view-invariant representation of a face but did not utilize the class information to separate images of different expressions in its pre-training stage.

\subsection{Contrastive Learning}

With the introduction of SimCLR \cite{simclr}, contrastive learning has recently emerged as an effective framework for representation learning from unlabeled data \cite{henaff2020data, hjelm2018learning, sermanet2018time, tian2020contrastive, wu2018unsupervised}. This framework is proven to be effective in learning useful representations in different domains, including FER \cite{li2022crs, xia2021micro, ST-CLR}. Contrastive learning can be utilized as a pre-training step to learn the underlying representation of the data in a self-supervised manner, followed by a fine-tuning step to adapt to the downstream task \cite{li2022crs}. This approach has also been shown to be useful in learning micro-expressions \cite{xia2021micro}. In \cite{fang2023rethinking}, contrastive learning was utilized for expression recognition in semi-supervised learning. Additionally, it has been utilized in video expression recognition \cite{ST-CLR}, where the model used contrastive learning with a novel temporal sampling technique to learn expression features in videos.

Contrastive learning methods generally learn from augmentations of input data by defining positive and negative samples. Here, the positive samples are different representations of the same image, while the negatives are representations of different images altogether. Hence, contrastive methods lend themselves well to self-supervised learning. 
The loss term used for contrastive learning is generally similar to triplet loss \cite{triplet}, where an anchor image representation is brought closer to the positive samples and moved away from the negatives. The key distinction between the triplet loss and contrastive loss is the number of positive and negative samples, which in the case of contrastive learning is one positive and multiple negatives as opposed to one positive versus one negative used with triplet loss. More recently, contrastive frameworks have even been shown to work well without the need for hard negative mining or memory banks \cite{misra2020self, wu2018unsupervised, he2020momentum}.

While contrastive learning is generally used to perform self-supervised training, it has also been shown to be advantageous in fully supervised settings \cite{supcon,zheng2021weakly}. For instance, contrastive learning has been used with class information to form positive samples in \cite{supcon}. In fact, it was mentioned in the paper that contrastive self-supervised pre-training can inadvertently push the images of the same class apart from one another, which is an undesired effect. To remedy this, the paper considered all the images from the same class, along with their augmentations as positive samples and all other images of different classes as negatives. This approach bears conceptual similarities to a few previous methods, such as the soft-nearest neighbours loss introduced in \cite{salakhutdinov2007learning}, which used Euclidean distance as a measure to separate instances of different classes. This was later improved in \cite{wu2018improving} by introducing the inner product as a distance measure. To learn better representations, a similar loss function was used in \cite{frosst2019analyzing} in the intermediate layers of the network rather than the final layer. 
COP-Gen \cite{li2022optimal} focused on generating optimal positive pairs for contrastive learning. 
PDNA \cite{miyai2023rethinking} proposed a new augmentation scheme for contrastive learning where a rotated image is considered either positive or negative based on semantic similarity. 
Contrastive learning has also been popular in video recognition tasks \cite{concur}, active learning \cite{roy2023active} and others.

\section{Method}\label{sec:method}

In this section, we describe the details of the proposed method. First, we present the problem setup followed by the proposed ViewFX method, Fine-tuning setup for downstream FER, network architecture, and the implementation details. 

\subsection{Problem Setup}
Let $D=\{(x_i, y_i)\}_{i=1..M}$ represent a training set of facial expression images, where $y_i \in\{1,2,...C\}$ is the corresponding class label and $C$ is the total number of expression classes. Here, images are captured from $v$ different angles, some of which may be challenging for FER (e.g. due to sharp camera angles). Our goal is to learn view-invariant features given the $v$ different views of an image, so that during inference, the model can predict the expression label from a \textit{single} image irrespective of its view and achieve high accuracy even on the challenging views.

\subsection{Proposed Method}
We propose ViewFX, a deep neural model that utilizes a new multi-term loss to learn strong view-invariant representations of expression from $D$, given the $v$ available views. There are three design considerations in our proposed method: (\textbf{1}) we aim to learn view-invariant representations of facial expressions; (\textbf{2}) we aim for the obtained view-invariant representations to be distinct for each expression class, allowing for strong FER; (\textbf{3}) we aim for the obtained representations to be `compact' and dis-ambiguous, allowing for even stronger FER. In the following subsections, we describe the components of our model that achieve each of these objectives. But first, we provide a brief preliminary discussion on contrastive learning, as it plays an important role in different components of our proposed method.

\subsubsection{Preliminaries}
The contrastive learning framework SimCLR \cite{simclr} learns useful data representations by maximizing the agreement between positive samples. Here, positive samples are different representations (augmentations) of a given input image. At the same time, it reduces the similarity of the input image with respect to negative samples (all other images). This form of contrastive learning is also known as instance discrimination since it considers only augmentations of the same image as positives.
 
For each image $x_i$, two augmentations are applied to generate two augmented images $\hat{x_i} = Aug(x_i)$. These are first passed through an encoder to obtain an embedding representation $r_i = Enc(\hat{x_i})$. Then, the encoder representation is passed through a shallow neural network called the projection head to obtain the final projection embedding vector $z_i = Proj(r_i)$. For a batch size of $N$, let $i\in\{1, ... 2N\}$ be the index of a randomly augmented image, and $\kappa(i)$ be the index of the second augmented image. The self-supervised contrastive learning loss is accordingly defined as:
\begin{equation}
\label{eq:loss_self}
    \mathcal{L}_{\text{\textit{self}}} = - \sum_{i=1}^{2N} log\frac{exp(z_i, z_{\kappa(i)}/\tau)}{\sum_{k=1}^{2N} \mathbf{1}_{[k \neq i]} exp(z_i, z_k/\tau)} ,
\end{equation}
where, $\mathbf{1}_{[k \neq i]}$ is an indicator function which returns 1 when $k$ is not equal to $i$, and 0 otherwise, and $\tau$ is a temperature parameter.

\subsubsection{View-invariant Representation Learning}
Through our proposed solution, we aim to learn view-invariant features of expression where similar representations are generated for an input image irrespective of its viewing angle. This effectively implies generating an embedding space where different views of the same image are situated close to one another. To achieve this objective, we propose a view-invariant contrastive loss that brings the representations for different views of an image close and pushes the cluster away from other images. The proposed view-invariant contrastive loss is derived from Equation \ref{eq:loss_self}, where instead of considering only the augmentations of the same image as positives (instance discrimination), we expend the definition of positive samples to all the views and augmentations of an input image. All other images are consequently considered negatives. This modification results in the following loss function: 
\begin{equation}
\label{eq_comp}
\begin{split}
    \mathcal{L}_{\text{\textit{view}}} = \sum_{i=1}^{2N} \bigg( \frac{-1}{2N_{\phi(i)}-1} \sum_{j=1}^{2N}
    \mathbf{1}_{i\neq j} \cdot \mathbf{1}_{\phi(i)=\phi(j)}
    \cdot \\
    log\frac{exp(z_i\cdot z_j/\tau)}{\sum_{k=1}^{2N} \mathbf{1}_{[k \neq i]} exp(z_i \cdot z_k/\tau) } \bigg)~,
\end{split}
\end{equation}
where, $\phi(i)$ is a view-invariant index for image $x_i$, which is constant across all the views of that image, and $2N_{\phi(i)} -1 $ is the number of images with the same view-invariant index.

\subsubsection{Expression Representation Learning}

Once view-invariant representations are effectively learned, we need to learn the correspondence between those representations and expression class labels. To do so, we utilize a supervised contrastive loss \cite{supcon} that maps the learned representations to expression labels. The supervised contrastive loss is also derived from Equation \ref{eq:loss_self}, where all the images of the same expression are considered positive samples while images of different expressions are considered negatives. This modification results in the following equation:
\begin{equation}
\label{eq:loss_con}
\begin{split}
    \mathcal{L}_{\text{\textit{sup}}} = \sum_{i=1}^{2N} \bigg( \frac{-1}{2N_{y_i}-1} \sum_{j=1}^{2N}
    \mathbf{1}_{i\neq j} \cdot \mathbf{1}_{y_i=y_j}
    \cdot \\
    log\frac{exp(z_i\cdot z_j/\tau)}{\sum_{k=1}^{2N} \mathbf{1}_{[k \neq i]} exp(z_i \cdot z_k/\tau) } \bigg)~,
\end{split}
\end{equation}
where, $y_i$ is the class label of instance $i$. Accordingly, the indicator function $\mathbf{1}_{y_i=y_j}$ returns 1 when index $i$ and $j$ are instances of the same class, and $2N_{y_i}-1 $ is the number of positive samples from class $y_i$. 

\begin{figure*}[t]
    \centering
    \includegraphics[width=0.85\linewidth]{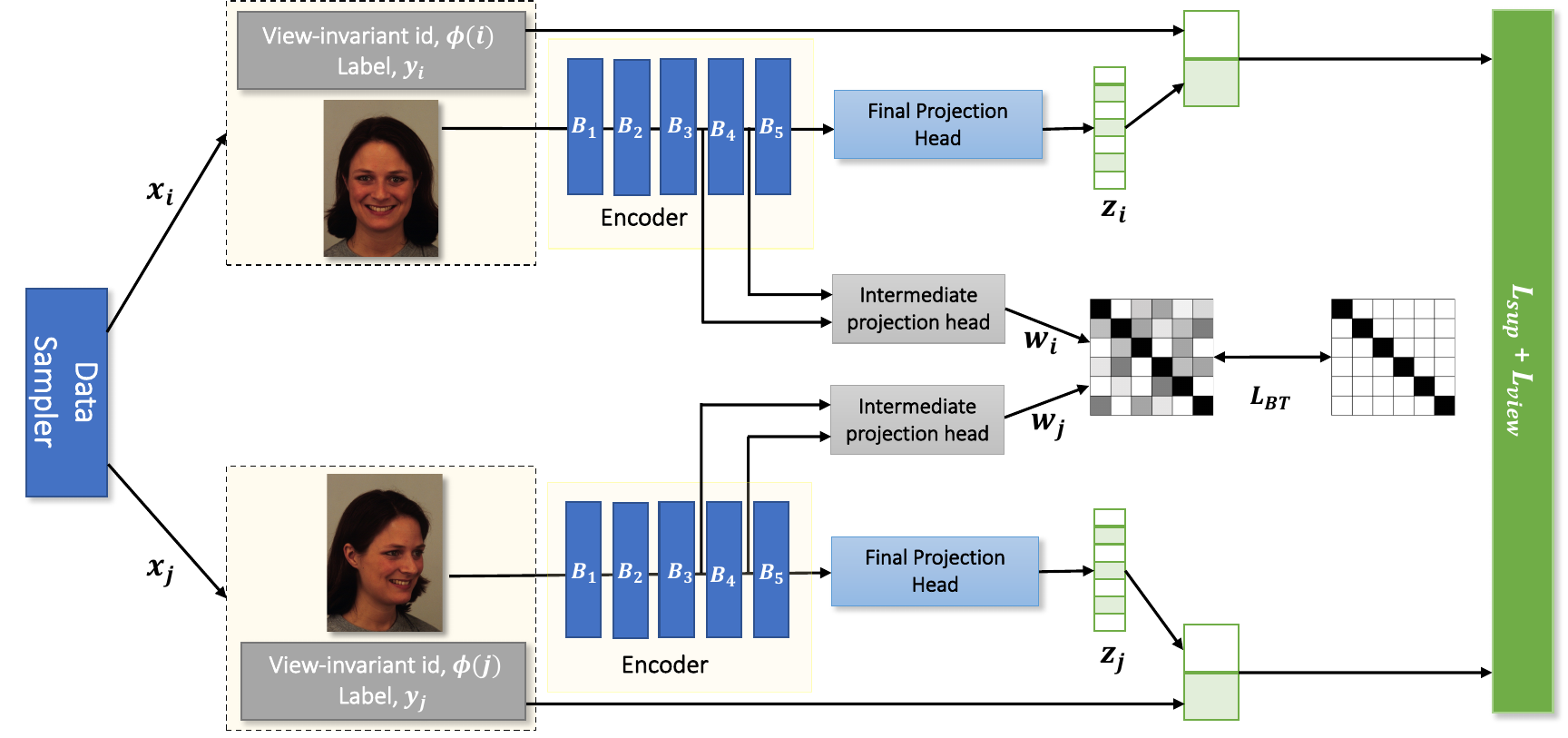} 
   \caption{Illustration of the proposed training framework. Here, $B_1,\cdots,B_5$ represent Block$_1$ to Block$_5$ in the proposed encoder. Intermediate outputs from $B_3$ and $B_4$ feed the two individual instances of the intermediate projection head. A final projection head generates the final output embedding on which the supervised contrastive loss and multi-view contrastive loss are applied.}
\label{fig:method}
\end{figure*}

\subsubsection{Compact Representation Learning}
Facial expressions are often distinguished by very subtle differences in the learned representations. Consequently, it is desired for FER solutions to learn representations that are more distinct and discriminative. Inspired by \cite{vicreg, barlowTwin}, where a new loss term was proposed to de-correlate learned representations, we incorporate the Barlow twins loss term in our encoder (see Figure \ref{fig:method}). This can help generate compact representations by measuring the correlation between the embedding vectors of the augmented images and forcing them to become as close to an identity matrix as possible.

Let's represent the embedding obtained from a given layer based on two augmentations as $w^A$ and $w^B$, respectively. Accordingly, the cross-correlation matrix between the two representations along the batch dimension is calculated as:
\begin{equation}
        C_{ij} = \frac{\sum^N_{b=1} w_{b,i}^A w_{b,j}^B}{ {\sqrt{\sum^N_{b=1} (w_{b,i}^A)^2} \sqrt{\sum^N_{b=1} (w_{b,j}^B)^2}}} , 
    \end{equation}
where $b$ is the batch index, and $i$ and $j$ are the vectors' dimension indices. Accordingly, the Barlow twins loss function is defined as:
    \begin{equation}
        \mathcal{L}_{BT} = \sum_i (1 - C_{ii})^2 + \alpha
        \sum_i \sum_{j \neq i} C_{ij}^2.
    \end{equation}
Here, the first term is called the `invariance' term, which tries to equate the diagonal elements of the cross-correlation matrix to 1, forcing it to learn representations that are invariant to distortion/augmentations applied to the image. The second term is called `redundancy reduction', which tries to equate the off-diagonal terms to 0, effectively de-correlating the different vector components of the representation. Accordingly, $\alpha$ is the coefficient of the second term, manually set between 0 and 1.

It has been shown in prior work \cite{frosst2019analyzing,inception} that using an auxiliary loss function to train the intermediate layers solely can boost gradient signals and further enable them to propagate backward, resulting in better overall training. Similarly, we find it more beneficial to utilize the $\mathcal{L}_{BT}$ in the intermediate layers rather than the entire network. As a result, we empirically find that utilizing $\mathcal{L}_{BT}$ on the two intermediate layers of the encoder yields the best results, which is what we use in our model. Our experiments in Section \ref{sec:ablation} further confirm that this design choice indeed boosts performance.

\subsubsection{Proposed Total Loss Function}
The proposed total loss function for our ViewFX method is the combination of the three loss functions mentioned above. Thereby, the final proposed loss can be represented as follows:
\begin{equation}
\label{final_eqn}
    \mathcal{L}_{ViewFX} =  \mathcal{L}_{sup} + \gamma * \mathcal{L}_{view} + \beta * (\mathcal{L}_{BT_1} + \mathcal{L}_{BT_2}),
\end{equation}
where $\gamma$ and $\beta$ are the coefficients of the multi-view and BT loss terms, respectively, and $\mathcal{L}_{BT_1}$ and $\mathcal{L}_{BT_2}$ are the losses used only to train the two intermediate layers of the encoder.

Figure \ref{fig:method} depicts the training pipeline of our proposed method. A data sampler samples a batch of images from the training set, and applies two augmentations yielding two sets of images $x_i$ and $x_j$. 
The sampled images are then passed through the encoder. The output from the encoder then goes through the projection head and outputs the projection embedding vector $z_i$ and $z_j$, on which the supervised contrastive loss and multi-view contrastive loss function are applied. The encoder outputs two other intermediate representations that are passed through the intermediate projection head to generate $w_i$ and $w_j$, on which the Barlow twins loss is applied.

\subsection{Fine-tuning for Downstream FER}
After pre-training the encoder with the proposed $\mathcal{L}_{ViewFX}$ loss, we fine-tune the model for the final FER task. Although the projection heads are useful for learning good representations in contrastive learning settings, it is no longer required for the downstream FER. As a result, we discard the projection heads and add a linear classification layer in place of the final projection head. We then fine-tune the model in a fully supervised setting with the ground-truth expression labels.

\begin{table}[t]
    \caption{Architectural details of the proposed model.}
    \label{tab_encoder}
    \begin{center}
    \small
    \setlength
    \tabcolsep{9pt}
    \begin{tabular}{c|c|c}
    \hline
    \textbf{Module} &\textbf{ Kernel }&\textbf{ Output Size} \\ 
    \hline 
    \multirow{2}{*}{\textit{$Block_1$}} &  7$\times7$, 64, \textit{s}=2  & \multirow{2}{*}{64$\times$112$\times$112}\\
    & \textit{max pool}, 3$\times$3, \textit{s}=2 \\
    \hline

     \multirow{3}{*}{\textit{$Block_2$}}&
     \blockb{256}{64}{3} & 
     \multirow{3}{*}{256$\times$56$\times$56} \\ 
    &&\\ 
    &&\\
    \hline
    
    \multirow{3}{*}{\textit{$Block_3$}}&\blockb{512}{128}{4}& \multirow{3}{*}{512$\times$28$\times$28} \\ 
    &&\\ 
    &&\\
    \hline
    
    \multirow{3}{*}{\textit{$Block_4$}}&\blockb{1024}{256}{6}& \multirow{3}{*}{1024$\times$14$\times$14} \\ 
    &&\\ 
    &&\\
    \hline
    
    \multirow{3}{*}{\textit{$Block_5$}}&\blockb{2048}{514}{3}& \multirow{3}{*}{2048$\times$7$\times$7} \\ 
    &&\\ 
    &&\\
    \hline
    
    \multirow{3}{*}{\textit{Final Proj.}} & \textit{Ada. Avg. Pool()} & 2048$\times$1$\times$1 \\
    & \textit{Reshape()} & 2048\\
     ($\textit{Block}_f$)& \textit{Linear()} & 512 \\
    & \textit{Linear()} & 128 \\
    \hline
    
    \multirow{2}{*}{\textit{$Int. ~Proj_1$}} & 3$\times$3, \textit{s}=2, \textit{p}=1 & 2048$\times$7$\times$7\\
    & $\textit{Block}_f()$ & 128\\
    \hline
    
    \multirow{3}{*}{\textit{$Int. Proj_2$}} & 3$\times$3, \textit{s}=2, \textit{p}=1 & 512$\times$14$\times$14\\
    & 3$\times$3, \textit{s}=2, \textit{p}=1 & 2048$\times$7$\times$7 \\
    & $\textit{Block}_f()$ & 128\\
    
    \hline
    \end{tabular}
    \end{center}
\end{table}

\subsection{Network Architecture}

\label{section_architecture}
We adopt ResNet-50 \cite{resnet} as the encoder for the proposed method, which consists of 5 main blocks. A final projection head (a shallow neural network with two fully connected layers) is added after the final convolution block of the ResNet-50 to generate the final output embeddings. In addition to the final projection head, two intermediate projection heads are attached to the end of the 3rd and 4th blocks of the ResNet-50. The intermediate projection heads are shallow neural networks consisting of a convolution layer, an adaptive average pooling layer, and two fully connected layers. The use of a multi-layer projection head is shown to be useful for learning good representations in some previous works \cite{simclr, supcon, barlowTwin}. The details of the modified ResNet-50 are presented in Table \ref{tab_encoder}.

\subsection{Implementation Details}
In this subsection, we describe the implementation and training details that are crucial to train the model. 

\begin{table}[!t]
    \caption{Parameters of augmentations.}
    \small
    \begin{center}
    \begin{tabular}{ l l }
    
    \toprule
    \textbf{Augmentation} & \textbf{Parameter} \\ \midrule 
    {Random Crop} & \texttt{0.2 to 1} \\ \hline
    {Horizontal Flip}  & \texttt{p = 0.5} \\ \hline
    {Color Jitter}     & \texttt{\specialcellleft{brightness = 0.4 to 1\\contrast = 0.4 to 1\\saturation = 0.4 to 1\\hue = 0.2 to 1}} \\ \hline
    {Gray Scale}       & \texttt{p = 0.2} \\ \hline
    {Gaussian Blur}    & \texttt{p = 0.5} \\
    \bottomrule
            
    \end{tabular}
    \label{tab_aug_value}
    \end{center}
\end{table}

\begin{table}[!t]
    \caption{Augmentations used in different stages of training.}
    \begin{center}
    \small
    \begin{tabular}{l|ccccc}
    \hline
    & \textbf{RC} & \textbf{HF} & \textbf{CJ} & \textbf{GS} & \textbf{GB} \\
    \hline 
    Pre-train & \cmark & \cmark & \cmark & \cmark & \cmark   \\
    Fine-tune &  \cmark & \cmark & \xmark & \xmark & \xmark   \\
    Test & \xmark & \xmark & \xmark & \xmark & \xmark \\
    \hline
    \end{tabular}
    \label{tab_stage_augmentations}
    \end{center}
\end{table}

\subsubsection{Augmentation Module}
The augmentation module, which generates the positive samples, is a very important component of contrastive training. The choice of augmentations is shown to be critical to the final performance of the model. The augmentations in our method are random resize crop (RC), random colour jitter (CJ), random horizontal flip (HF), random Gaussian blurring (GB), and random gray-scaling (GS). The augmentation module applies random flip and random blurring with a probability of 0.5 and random gray scaling with a probability of 0.2. For random cropping, a crop area between 0.2 and 1 is chosen. For colour jittering, the brightness, saturation, and contrast are randomly altered with a coefficient randomly chosen between 0.4 and 1, while Hue is altered with a coefficient randomly chosen between 0.2 and 1. The parameters for different augmentations are summarized in Table \ref{tab_aug_value}. An illustration of augmented images using our augmentation module is shown in Figure \ref{fig:aug}.

Although hard augmentations are helpful in the contrastive pre-training stage, the fine-tuning stage uses fewer augmentations. The augmentations used in different stages of training are summarized in Table  \ref{tab_stage_augmentations}. For the pre-train stage, we use all the augmentations mentioned above. In the fine-tuning stage, we only used random resize crop and horizontal flip. Finally, we test the model using no augmentations at all.

\begin{figure}
\input{figures/augs}
\end{figure}

\subsubsection{Training and Implementation Setups}
We pre-train the encoder with the proposed total loss for 1000 epochs using an Adam optimizer and a learning rate of 1e-3. In this step, we use a cosine learning decay over the training epochs, and weight decay of 1e-4. Once the pre-training step is done, we fine-tune the model for 75 epochs using a learning rate of 1e-4 and plateau learning rate decay, with a decay factor of 0.5 and patience of 3. Table \ref{tab_hyperparameters} summarizes the parameters used to pre-train and fine-tune the encoder. 
The proposed method is implemented with PyTorch and trained on 4 NVIDIA V100 GPUs. A PyTorch-like pseudo-code is illustrated in Algorithm \ref{algo:viewfx}.

\begin{table}[!t]
\caption{Hyper-parameter settings for different training stages.}
\begin{center}
\small
\begin{tabular}{lll}
\toprule
\textbf{Parameter} & \textbf{Pre-train}  & \textbf{Fine-tune} \\ \toprule
Input size & $224\times224$ & $224\times224 $\\
Training epochs & 1000 & 75\\
Optimizer  & Adam & Adam\\
Initial learning rate & 0.001 & 0.0001\\
Weight decay & 1e-4 & 1e-5 \\
LR Scheduler & Cosine & Plateau\\
Warm-up epochs & 10 & 0 \\
\bottomrule

\end{tabular}
\label{tab_hyperparameters}
\end{center}
\end{table}

\begin{algorithm}
    \caption{ViewFX algorithm}
    \label{algo:viewfx}
    \begin{algorithmic}[1] 
    \STATE Input: image, view\_id, label
    
    \FOR{layer $i = 1, 2 \;$}
        \STATE $x_i$= Aug(x)
        \STATE $r_{ia}$, $r_{ib}$, $r_{if}$ = $Enc(x_i)$
        \STATE $w_{ia}$, $w_{ib}$, $z_{i}$ =  $Proj_1(r_{ia})$, $Proj_2(r_{ib})$, $Proj_f(r_{if})$
    \ENDFOR 
    \STATE $L_1 = \mathcal{L}_{\text{\textit{sup}}}(z_{1}, z_{2}, label)$
    \STATE $L_2 = \mathcal{L}_{\text{\textit{view}}}(z_{1}, z_{2}, view\_id)$
    \STATE $L_3 = \mathcal{L}_{\text{\textit{BT}}}(w_{1a}, w_{2a})$
    \STATE $L_4 = \mathcal{L}_{\text{\textit{BT}}}(w_{1b}, w_{2b})$
    \STATE $Loss_{total} = L_1 + \gamma * L_2 + \beta * (L_3+L_4)$
    \end{algorithmic}
\end{algorithm}

\section{Experiments and Results}\label{results}

In this section, we describe the experiments and results of our proposed method on two public multi-view expression datasets, KDEF and DDCF. 

\subsection{Datasets}
Given our main goal of developing a model capable of learning strong \textit{view-invariant} representations, we need \textit{multi-view FER datasets} to test our solution. Among the available multi-view FER datasets, excluding 3D face datasets (which require the addition of further modules to the model), KDEF \cite{kdef}, DDCF \cite{ddcf}  and BU3DFE \cite{bu3d} are three large datasets that are publicly available. Following, we describe these three datasets in detail.

\subsubsection{KDEF \cite{kdef}} 
This is a multi-view dataset for facial expressions collected from 140 subjects. This dataset contains seven classes where each image is captured from 5 different camera angles. The views are abbreviated as +90$^{\circ}$: full right (FR), +45$^{\circ}$: half right (HR), 0$^{\circ}$: straight (S), -45$^{\circ}$: half left (HL), and -90$^{\circ}$: full left (FL).  

\subsubsection{DDCF \cite{ddcf}} 
This is another multi-view facial expression dataset which is collected from 80 subjects. This dataset is captured from 5 different camera angles: +60$^{\circ}$: full right (FR)), +30$^{\circ}$: half right (HR), 0$^{\circ}$: straight (S), -30$^{\circ}$: half left (HL), and -60$^{\circ}$: full left (FL). The DDCF dataset contains eight facial expression classes.

\subsubsection{BU3DFE \cite{bu3d}}
This dataset is another multi-view dataset collected from 100 subjects (56\% female and 44\% male) with different age groups and ethnic/racial backgrounds. The images captured from -45$^{\circ}$ to  +45$^{\circ}$. This dataset contains six facial expressions: happiness, disgust, fear, anger, surprise and sadness

\subsection{Evaluation} 
Accuracy is measured for the downstream network, which predicts the output expression labels. The input to this network is a single image (a single view of a subject performing an expression), as the goal of this study is to perform FER (at inference time) for an image captured from any view. 
The accuracy is therefore measured as the rate of correct classifications over all the test images.

\subsection{Results}
\label{sec_results}
We perform various experiments to fully evaluate the performance of ViewFX with respect to different view angles, and its performance against different expressions. We also perform experiments on different amounts of labelled data, hyper-parameter choices, and different encoders. Finally, we present a detailed comparison with prior works on these datasets.

\begin{figure}[!t]
  \centering
  \includegraphics[width=0.6\linewidth]{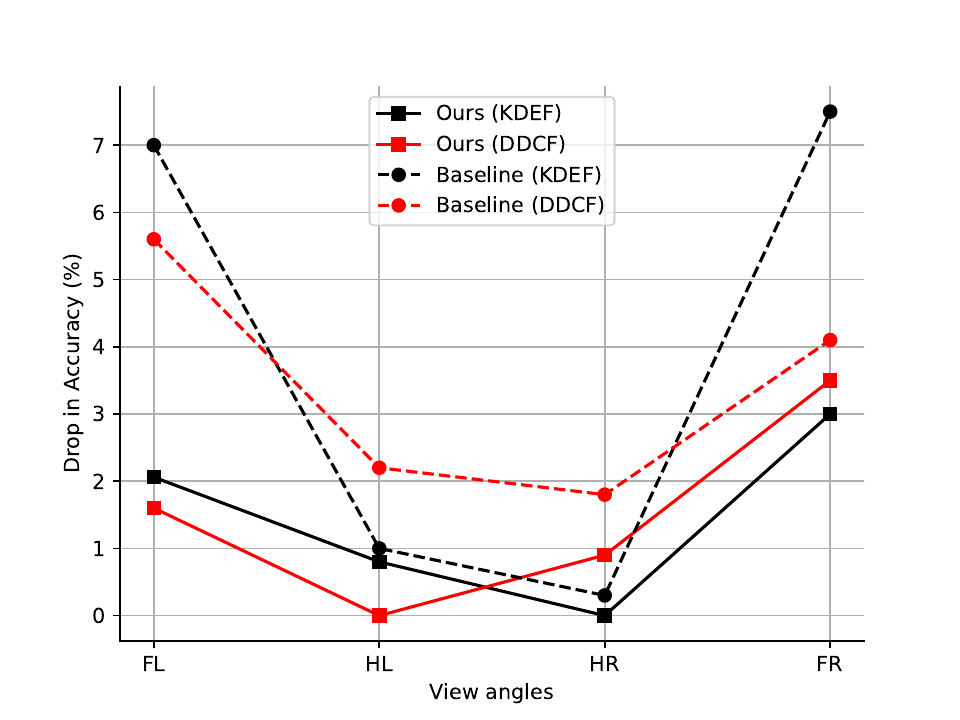}
  \caption{Drop in performance for different viewing angles with respect to the frontal view.}
  \label{fig:acc_diff}
\end{figure}

\begin{table}

\caption{Performance breakdown and comparison for different viewing angles for the KDEF dataset.}
\label{tab_angle}
\small
\begin{center}
\begin{tabular}{l | c c c c c}
\hline
\multirow{2}{*}{\textbf{Methods}} & \multicolumn{5}{c}{\textbf{Accuracy/Viwes}}  \\ \cline{2-6}
& \textbf{FL} & \textbf{HL} & \textbf{F} & \textbf{HR} & \textbf{FR}\\
\hline 
Cross-Entropy & 86.52 & 92.54 & 93.5 & 93.15 & 86.0 \\
PhaNet \cite{PhaNet} & 90.2 & 90.2 & 89.7 & 83.5 & 89.1\\
LDBP \cite{santra2016local} & - & 80.76 & 83.51 & 80.22 & - \\
CL-MEx \cite{CL_MEx} & 93.57 & 96.11 & 96.32 & 96.21 & 91.0 \\

\textbf{ViewFX} & 96.49 & 96.65 & 98.55 & 98.55 & 95.55 \\
\hline
\end{tabular}
\end{center}
\end{table}

\subsubsection{Sensitivity to Different Views}
A key goal of the proposed method is to be able to achieve robust performance against every view, especially more challenging ones. In this regard, frontal views are often the easiest since the entire face is visible, while the HL and HR views are somewhat more challenging. Accordingly, the FL and FR views are the most difficult since only one side of the face is visible. Table \ref{tab_angle} presents the FER results separated for different angles on the KDER dataset, and its comparison with prior works that have reported such a breakdown (no prior work reported this breakdown for DDCF). We observe from the table that the proposed method performs better across all views. We also observe that for the most challenging angles (FL and FR), our method performs considerably better than prior works.

\begin{figure}[!t]
    \centering
    \begin{subfigure}[t]{0.45\columnwidth}
        \centering
        \includegraphics[height=1.8in]{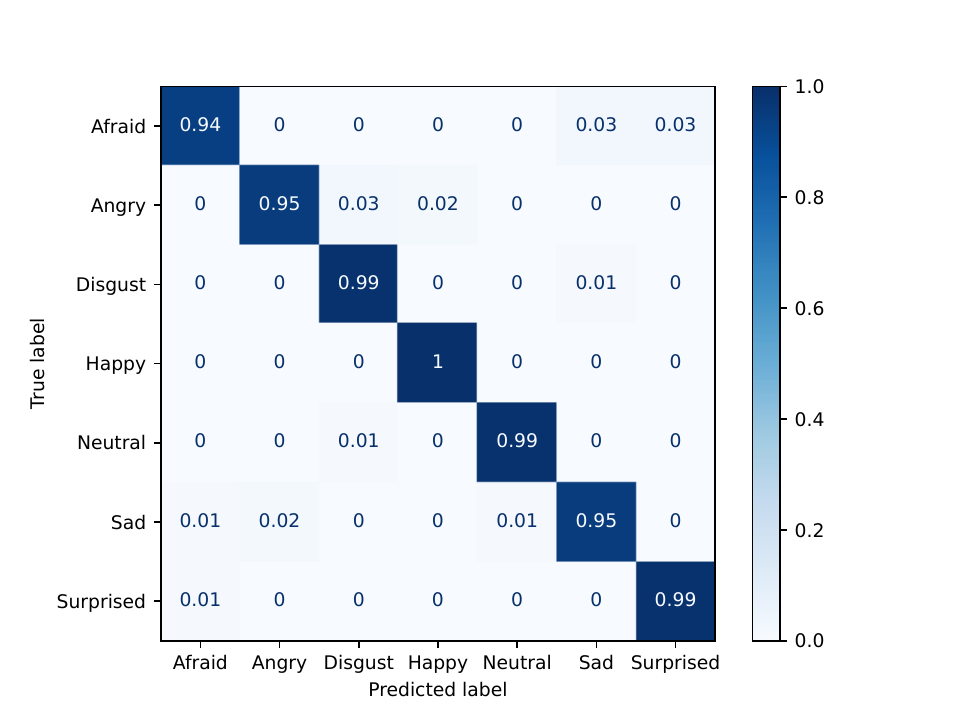}
        \caption{KDEF}
    \end{subfigure}
    \begin{subfigure}[t]{0.45\columnwidth}
        \centering
        \includegraphics[height=1.8in]{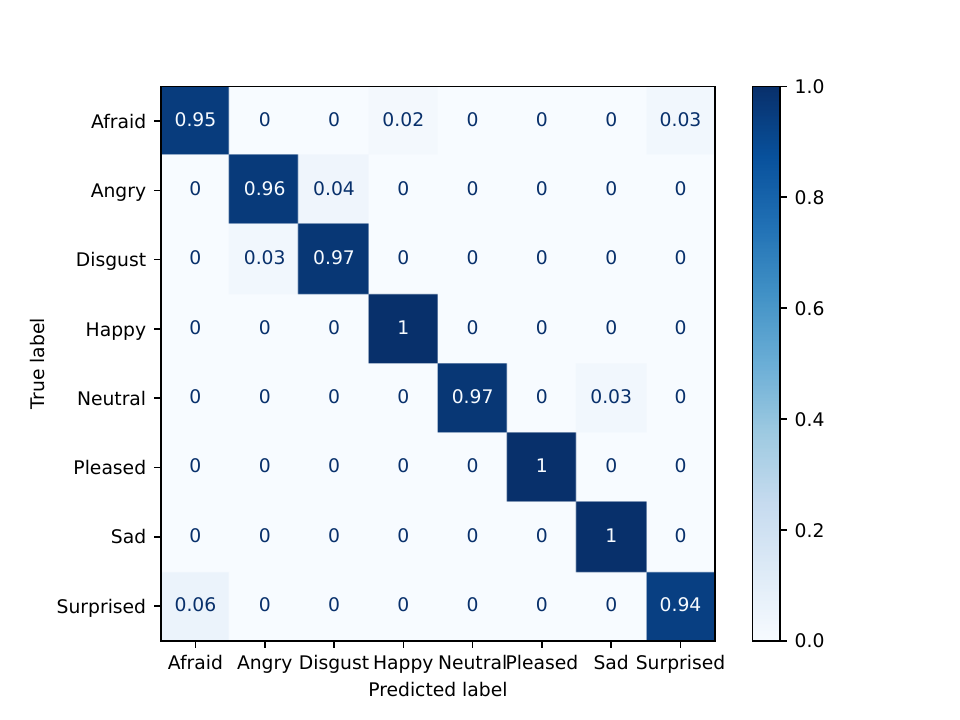}
        \caption{DDCF}
    \end{subfigure}
    \caption{Confusion matrices for different datasets.}
    \label{fig:confusion}
\end{figure}

In Figure \ref{fig:acc_diff} we show the \textit{drop} in accuracy for different views \textit{compared to the frontal} (easiest view) view for: (\textit{i}) ViewFX, (\textit{ii}) the same encoder when a standard cross-entropy loss is used for training (baseline). As demonstrated, there is a considerable drop in performance in FL and FR views (most challenging views) for the baseline (approximately 7\% drop for KDEF, and 4-5.5\% drop for DDCF). For HL and HR views (somewhat challenging views), drops of around 2\% and 1\% are observed by the baseline model for KDEF and DDCF. We observe, however, that ViewFX shows considerably less sensitivity towards the difficult views as the performance of ViewFX doesn't exhibit strong degradation for the challenging views (only 1.5-3.5\% drop) in comparison to the frontal view.

\subsubsection{Sensitivity to Different Expressions}
Figure \ref{fig:confusion} presents the breakdown of our results for different expressions. The illustrated confusion matrices clearly show that our model performs well across all expressions for both datasets, with the main mistakes being made between `Disgust' and `Angry', as well as between `Surprised' and `Afraid', which are inherently challenging expressions to distinguish.

\begin{figure}[!t]
  \centering
  \includegraphics[width=0.6\linewidth]{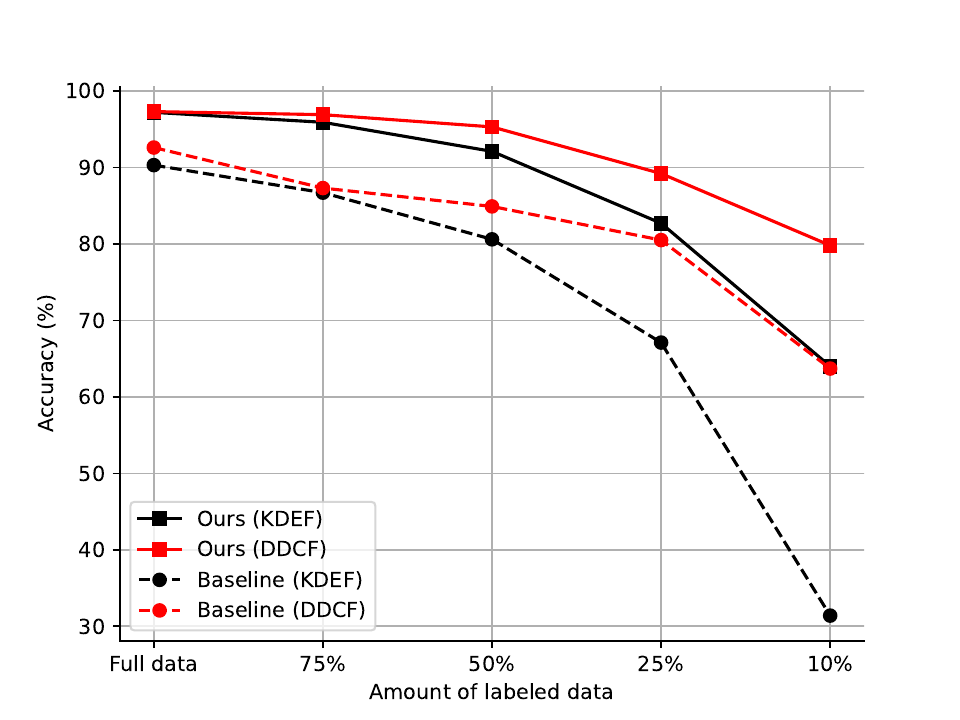}
  \caption{Performance for different amounts of output labels used in training.}
  \label{fig:partial_data}
\end{figure}

\subsubsection{Sensitivity to the Amount of Labelled data}

Next, we evaluate the sensitivity of our method against the amount of labelled data used in the training. Figure \ref{fig:partial_data} presents the performance of our method in comparison to the baseline (cross-entropy loss) when the amount of labelled data is reduced. We observe that ViewFX maintains relatively steady performance when the amount of labelled data is reduced to 75\%, 50\%, 25\%, and even 10\%. However, the accuracy of the baseline model drops significantly with the reduced labelled data. For instance, in the extreme case of using only 10\% of the labels, the drops in accuracy are approximately 12\% and 30\% for our method in KDEF and DDCF, in comparison to the baseline's respective 25\% and 60\% drops, highlighting considerably less sensitivity to labelled data by our method.

\begin{table}[!t]
\caption{Performance for different sizes of intermediate (\textit{in}) and final (\textit{fi}) projection embeddings.}
\label{tab:projection}
\small
\setlength
\tabcolsep{10pt}
\begin{center}
\begin{tabular}{c |  c | c c c}
\hline
Dataset & & \textit{fi=}\textbf{128} & \textit{fi=}256 & \textit{fi=}512 \\
\hline
\multirow{3}{3em}{KDEF}& \textit{in=}128 & 96.13 &  96.74 &  96.54 \\
& \textit{in=}\textbf{256} & \textbf{97.15} &  97.05 &  96.95 \\
& \textit{in=}512 & 96.13 &  97.05 &  96.74 \\
\hline
\multirow{3}{3em}{DDCF}& \textit{in=}\textbf{128} &\textbf{ 97.34} &  97.15 &  96.94 \\
& \textit{in=}256 & 97.30 &  96.82 &  96.73 \\
& \textit{in=}512 & 96.58 &  96.32 &  96.13 \\
\hline
\end{tabular}
\end{center}
\end{table}

\begin{table}[!t]
    \caption{Projection head configuration.}
    \begin{center}
    \small
    \begin{tabular}{cll}
    \toprule
    
    \textbf{Batch Norm} & \textbf{Linear layers} & \textbf{Accuracy} \\
    \hline
    \multirow{3}{*}{yes} & 512-128 & \textbf{97.15\%}\\
    & 1024-128 & 96.97\%\\
    & 512-512-128 & 96.52\%\\ \hline
    \multirow{3}{*}{no} & 512-128 & 96.92\%\\
    & 1024-128 & 96.87\%\\
    & 512-512-128 & 96.12\%\\
    
    \bottomrule

    \end{tabular}
    \label{tab_linear}
    \end{center}
\end{table}

\begin{table}[!t]
\caption{Performance for different values of $\gamma$ and $\beta$.}
\small
\begin{center}
\begin{tabular}{c | c c c c c}
\hline

 &$\gamma=$ 0.1 & $\gamma=$ 0.25 & $\gamma=$ \textbf{0.5} & $\gamma=$ 0.9 & $\gamma=$ 1.0 \\
\hline
KDEF & 95.32 & 96.74 &\textbf{ 97.15} &  95.11  & 94.3 \\
DDCF & 96.19 & 96.58 & \textbf{97.34} &  97.11  & 97.01 \\
\hline \hline
 & $\beta=$ 0.1 & $\beta=$ 0.25 & $\beta=$ 0.5 & $\beta=$ \textbf{0.9} & $\beta=$ 1.0 \\
\hline
KDEF & 95.72 & 96.13 & 96.54& \textbf{97.15} & 96.95\\
DDEF & 96.87 & 96.97 & 97.17 & \textbf{97.34} & 97.30 \\
\hline
\end{tabular}
\label{tab_gamma}
\end{center}
\end{table}

\subsubsection{Sensitivity to Model Settings}

Here, we study the sensitivity of the proposed model toward different choices of model settings and hyper-parameters. As mentioned previously, the choice of projection head dimensions can have a significant impact on the downstream performance of the model. We, therefore, conduct sensitivity analyses with different values for the dimensions of both the intermediate and output projection heads. The results in Table \ref{tab:projection} show that the proposed method is quite stable against the choice of embedding dimensions. The best results are achieved with the final dimension of $128$ for both datasets and intermediate dimensions of $128$ and $256$ for DDCF and KDEF, respectively. 

We have also conducted experiments with different numbers of linear layers and using Batch Normalization. The results are summarized in Table \ref{tab_linear}. We obtain the best result for a 2-layer projection head and an embedding dimension of 512-128. Scaling up the number of layers or the dimension of the linear layer hurts the final accuracy of the model. We also find that Batch Normalization is important for achieving the best results.
\begin{figure}[!t]
  \centering
  \includegraphics[width=0.6\linewidth]{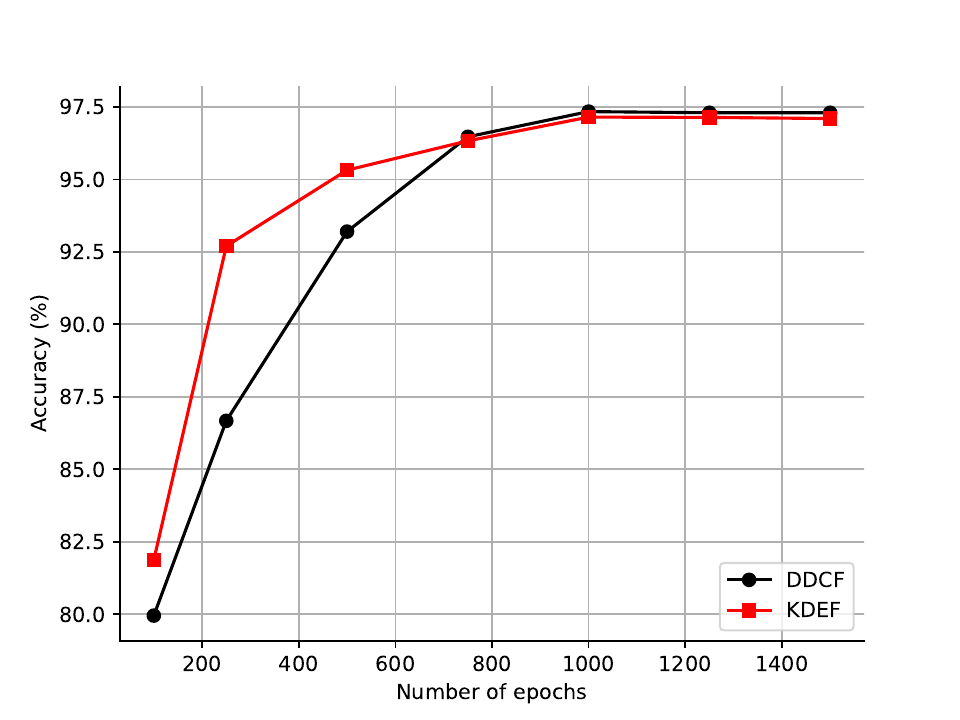}
  \caption{Performance for different number of epochs.}
  \label{fig:epochs}
\end{figure}

According to Eq. \ref{final_eqn}, our proposed total loss consists of 3 loss terms with 3 unique responsibilities. To balance the impact of each term, two hyper-parameters ($\gamma$ and $\beta$) were introduced. Here we perform experiments to evaluate the impact of these parameters on the performance. The results are presented in Table \ref{tab_gamma}, which shows that the best results are achieved for $\gamma=$ 0.5 and $\beta=$ 0.9 for both of the datasets.

Next, we experiment with the number of epochs for the pre-training, which is known to be an important hyper-parameter in contrastive settings. The results are shown in Figure \ref{fig:epochs}. As we see from the figure, the downstream accuracy of the model increases significantly with an increase in the number of epochs for up to 1000 epochs. After that, no further improvements are observed.

Finally, we show the importance of each of the augmentations used for pre-training the model in Table \ref{tab_aug_ablation}. In each experiment, we remove one of the augmentation modules used in the final method and observe the drop in the performance of the final accuracy after fine-tuning. We observe the highest amount of drop in accuracy with the removal of the colour-jittering augmentation module. A close performance drop is observed when the multi-scale cropping is removed. Both augmentations show a drop of over 4\% in accuracy. Horizontal Flip also has an impact of close to 3\%. A comparatively lower drop of around 1\% is seen when the gray scaling and Gaussian blur augmentations are removed. 

\begin{table}[!t]
\caption{Sensitivity study on the effect of different augmentations conducted on the KDEF dataset.}
\begin{center}
\small
\begin{tabular}{ccccc|c}
\hline
\multicolumn{5}{c}{\textbf{Augmentations}} & \multicolumn{1}{c}{\textbf{Accuracy}}\\ \hline
 \textbf{MSC} & \textbf{HF} & \textbf{CJ} & \textbf{GS} & \textbf{GB} & \textbf{KDEF} \\
\hline 
 \cmark & \cmark & \cmark & \cmark & \cmark &  97.15\%\\
\xmark & \cmark & \cmark & \cmark & \cmark &  92.85\%\\
 \cmark & \xmark & \cmark & \cmark & \cmark &  94.69\%\\
 \cmark & \cmark & \xmark & \cmark & \cmark &  92.45\% \\
 \cmark & \cmark & \cmark & \xmark & \cmark &  96.13\%\\
  \cmark & \cmark & \cmark &  \cmark & \xmark &  96.01\%\\
\hline
\end{tabular}
\label{tab_aug_ablation}
\end{center}
\end{table}

\begin{table*}[!t]
\caption{Accuracy of the proposed method for different encoders for KDEF and DDEF dataset.}
\definecolor{rc1}{RGB}{235,235,235}
\definecolor{rc2}{RGB}{255,255,255}
\small

\setlength
\tabcolsep{10pt}
\begin{center}
\begin{tabular}{c | c c c c c}
\hline
\textbf{Encoder} &\textbf{ \#param }&\textbf{ Dataset }& \textbf{Fully-Supervised }& \textbf{ViewFX} \\
\hline 

 \rowcolor{rc1}
 & & KDEF & 88.69\%& 95.41\% \\
  \rowcolor{rc1}
 \multirow{-2}{*}{ResNet-18} &\multirow{-2}{*}{11.7 M} &DDCF & 91.25\% & 95.66\%\\

 & & KDEF & 90.34\% & 96.50\%\\
 \multirow{-2}{*}{ResNet-50} &\multirow{-2}{*}{25.6 M} &DDCF & 91.56\% & 96.86\% \\

 \rowcolor{rc1}
 & & KDEF & 43.26\% & 95.32\% \\
  \rowcolor{rc1}
 \multirow{-2}{*}{ResNet-101} &\multirow{-2}{*}{44.5 M} &DDCF & 54.09\% & 94.29\%\\

 & & KDEF &64.08\% & 85.68\%\\
 \multirow{-2}{*}{VGG-16} &\multirow{-2}{*}{138.4 M} &DDCF & 67.97\% & 84.57\%\\

 \rowcolor{rc1}
 & & KDEF & 64.89\% & 87.25\%\\
  \rowcolor{rc1}
 \multirow{-2}{*}{VGG-19} &\multirow{-2}{*}{143.7 M} &DDCF& 68.72\%& 86.96\%\\

 & & KDEF & 71.63\% & 89.06\%\\
 \multirow{-2}{*}{AlexNet} &\multirow{-2}{*}{61.1 M} & DDCF & 75.25\%& 89.56\%\\

\hline
\end{tabular}
\end{center}
\label{tab_encoders}
\end{table*}

\subsubsection{Impact of Encoder}

In this study, we also experiment with a few other architectures as the encoder, including AlexNet \cite{krizhevsky2012imagenet} and VGG \cite{simonyan2014very}. More specifically, ResNet-18, ResNet-50, ResNet-101, VGG-16, VGG-19, and AlexNet encoders are explored in this study. For these experiments, pre-training was carried out for 500 epochs. Table \ref{tab_encoders} represents the accuracy of the proposed method as well as the fully supervised method with different encoders along with the number of parameters. As we observe from the table, for comparatively smaller models like ResNet-18 and ResNet-50, which have 11.7 and 25.6 million parameters, respectively, the supervised learning method exhibits a relatively low difference compared to ViewFX. In these cases, the ViewFX still shows around 4-6\% improvement in performance. However, this difference increases significantly when the parameter size of the model increases. For larger models, we hypothesize that fully supervised training results in overfitting, while this does not occur in our contrastive solution due to the effective use of augmentations. Although the proposed method shows improvement over the supervised setting for larger models, the best accuracy is achieved for the ResNet-50 encoder for both of the datasets.

\begin{table*}[!t]

\caption{Ablation experiments on the different components of the proposed framework. Here, `\textit{inter}' denotes the BT loss used in the intermediate layers, while `\textit{final}' denotes the BT loss used at the final layer (for comparison purposes). The results are presented for when 100\% and 50\% output labels are used for training.} 
\label{tab_ablation}
\setlength
\tabcolsep{10pt}
\small
\begin{center}
\begin{tabular}{l | l | l | l |l}
\hline
\multicolumn{1}{c|}{\textbf{Settings}} & \multicolumn{2}{c|}{\textbf{KDEF}} & \multicolumn{2}{c}{\textbf{DDCF}} \\
\cline{2-5}
\multicolumn{1}{c|}{\textit{Percentage of labeled data}}   & \textit{100\%} & \textit{50\%}& \textit{100\%} & \textit{50\%} \\
\hline

\textbf{$\mathcal{L}_{sup} +\mathcal{L}_{view} +\mathcal{L}_{BT} $} &\textbf{97.15} & \textbf{92.06} & \textbf{97.34} & \textbf{95.30}\\

Fully-supervised training & 90.34 & 80.61  & 92.58 & 84.92 \\ \hline


$\mathcal{L}_{sup}  +\mathcal{L}_{BT} (inter.)$  & 96.54 & 90.02 & 96.55&94.20\\
$\mathcal{L}_{view} +\mathcal{L}_{BT} (inter.)$ & 94.80&84.93 & 94.72&90.44\\
$\mathcal{L}_{sup}  +\mathcal{L}_{BT} (final)$  & 96.33 & 88.75 & 95.81&90.25 \\
$\mathcal{L}_{view} + \mathcal{L}_{BT} (final)$  & 94.68&81.53 & 94.62&86.98 \\
$\mathcal{L}_{sup} +\mathcal{L}_{view}$ & 96.95&90.84 & 96.87&93.26\\

{$\mathcal{L}_{sup}$} &  {95.17} & {89.14} &  {96.01} & {92.67}\\ 
{$\mathcal{L}_{view}$} & {94.71} & {87.56} & {94.96} & {90.07}\\ 
{$\mathcal{L}_{BT}$} & {92.34} & {86.23} & {93.59} & {88.56} \\ 
%
\hline
\end{tabular}
\end{center}
\end{table*}

\subsubsection{Ablation Experiments}\label{sec:ablation}

To understand the impact of different components of our proposed method, we conduct detailed ablation studies by systematically removing different parts of the model. 
We perform these experiments twice, once with all the labels used in training, and once when only 50\% of the output labels are used. 
Table \ref{tab_ablation} shows the results when different terms are removed from the proposed loss $\mathcal{L}_{ViewFX}$. As we observe, the proposed method shows a significant accuracy drop of nearly 6.8\% and 4.8\% for KDEF and DDCF datasets, respectively, when the $\mathcal{L}_{ViewFX}$ is totally removed (the model is trained in a fully supervised setting). 
Next, we remove the multi-view contrastive component from $\mathcal{L}_{ViewFX}$, followed by the removal of the supervised contrastive component from $\mathcal{L}_{ViewFX}$, where in both cases, we observe a drop in performance. Next, we repeat this experiment with the exception of using the $\mathcal{L}_{BT}$ for training the \textit{last} projection head instead of the intermediate projection heads, which results in relatively lower accuracy. This experiment validates the choice of using the $\mathcal{L}_{BT}$ for the intermediate layers as opposed to the final projection head. Lastly, we remove the $\mathcal{L}_{BT}$ term altogether and only use the supervised and multi-view contrastive losses, demonstrating the positive impact of the $\mathcal{L}_{BT}$. This experiment further highlights the added value of our method when smaller amounts of labels are available for training. 
Next, we remove two of the loss components at a time, effectively using a single loss at each setting. These experiments show the individual importance of each of the loss components. Here, we overserve the highest individual impact for the $\mathcal{L}_{sup}$, followed by $\mathcal{L}_{view}$.

\begin{table*}[!t]
\caption{Performance and comparison to previous works on KDEF and DDCF datasets.}
\definecolor{rc1}{RGB}{235,235,235}
\definecolor{rc2}{RGB}{255,255,255}
\small
\begin{center}
\begin{tabular}{ l| l| c |c |c | c| l}
\hline
        
         \textbf{Dataset} & \textbf{Method} & \textbf{Backbone} & No. of Param. & \textbf{Input Size}  & \textbf{Evaluation} &\textbf{Acc. $\pm$ SD}\\
        \hline\hline
        
        \multirow{8}{2em}{KDEF} &  SVM \cite{SVM} & SVM & - & $320\times240$ & 80–20\% split & 70.5$\pm$1.2\\
        & SURF \cite{SURF} & - & - & $40\times40$& 10-fold split& 74.05$\pm$0.9\\
         & TLCNN \cite{TLCNN} & AlexNet\cite{krizhevsky2012imagenet} & 60M & $227\times227$ & 10-fold split  & 86.43$\pm$1.0\\
         & PhaNet \cite{PhaNet} & DenseNet \cite{huang2017densenet} & 15.3M & $224\times224$& 10-fold split &  86.5\\
         & MPCNN \cite{MPCNN} & MPCNN \cite{MPCNN} & - & Multi-scale & 5-fold split & 86.9$\pm$0.6\\
        & RBFNN \cite {RB-FNN} & - & - & $256\times256$ & - & 88.87\\
        & CL-MEx \cite{CL_MEx} &  ResNet-18 & 11.7M &  $224\times224$& 10-fold split &94.64$\pm$0.92\\
        & CL-MEx \cite{CL_MEx} & ResNet-50 & 25.6M & $224\times224$  & 10-fold split & 94.71 $\pm$1.06\\

        \cline{2-7}
        \rowcolor{rc1}
        & \textbf{ViewFX }&  ResNet-18 & 11.7M &  $224\times224$& 10-fold split & \textbf{95.48$\pm$0.51}\\
        \rowcolor{rc1}
        & \textbf{ViewFX} &  ResNet-50 & 25.6M & $224\times224$& 10-fold split& \textbf{97.15$\pm$0.46}\\
        \hline
        
        \multirow{5}{2em}{DDCF} & LBP \cite {khan2019saliency} & - & - & $64\times64$ & 10-fold split &  82.3\\
        & SVM \cite {albu2015neural}& SVM &- & -&- & 91.27 \\
        & RBFNN \cite{albu2015neural}& RBF-NN & - & -&- & 91.80\\
        & CL-MEx \cite{CL_MEx} & ResNet-18 & 11.7M &  $224\times224$& 10-fold split & 95.26$\pm$0.84\\
        & CL-MEx \cite{CL_MEx} & ResNet-50& 25.6M & $224\times224$  & 10-fold split & 94.96 $\pm$0.92  \\
        
        \cline{2-7}
        \rowcolor{rc1}
        &\textbf{ViewFX} &  ResNet-18 & 11.7M &  $224\times224$& 10-fold split &\textbf{95.96$\pm$0.68}\\
        \rowcolor{rc1}
        &\textbf{ViewFX} & ResNet-50 & 25.6M & $224\times224$& 10-fold split & \textbf{97.34$\pm$0.54}\\
        \hline

        \multirow{9}{4em}{BU-3DFE}
        
        & GLF \cite{jie2020multi}& - & - & -&- & 81.5\\
        & MVFE-L \cite{jie2020multi}& sCNN & - & -&- & 88.7\\
        & PCA-LDA\cite {venkatesh2012simultaneous}
        & HNN & - & - & 
        - &  93.7\\
        & Huynh et al. \cite {huynh2016convolutional} & - &- & -&- & 92.0 \\
        &  Hariri et al. \cite {hariri20173d} & SVM &- & -&- & 92.62 \\
        & CL-MEx \cite{CL_MEx} & ResNet-18 & 11.7M &  $224\times224$& 10-fold split & 93.15$\pm$0.65\\
        & CL-MEx \cite{CL_MEx} & ResNet-50& 25.6M & $224\times224$  & 10-fold split & 94.25 $\pm$0.51  \\
        \cline{2-7}
        \rowcolor{rc1}
        & \textbf{ViewFX} &  ResNet-18 & 11.7M &  $224\times224$& 10-fold split &\textbf{96.04$\pm$0.56}\\
        \rowcolor{rc1}
        &\textbf{ViewFX} & ResNet-50 & 25.6M & $224\times224$& 10-fold split & \textbf{97.02$\pm$0.66}\\
        \hline

\end{tabular}
\label{tab_compare}
\end{center}
\end{table*}

\subsubsection{Comparison to Existing Methods.}
Table \ref{tab_compare} presents the performance of our method in comparison with prior works in the area. The backbone model, the number of parameters in the model, the resolution of input images, and the evaluation protocols are also presented in this table. We observe from the table that ViewFX achieves over 2.1\% improvement for the DDCF dataset, 2.4\% for the KDEF dataset, and 2.67\% for the BU-3DFE dataset, even though some of the previous methods used higher input resolutions and larger models. For a fair comparison with CL-MEx, which used ResNet-18 as an encoder, we report the accuracy of ViewFX with the ResNet-18 encoder as well as CL-MEx trained on ResNet-50. The results show that CL-MEx does not exhibit much improvement with ResNet-50 on KDEF, and shows a decline in accuracy on DDCF. However, the proposed ViewFX shows considerable improvements over CL-MEx when ResNet-50 is used.

\section{Conclusion}

This paper introduces a contrastive learning framework for \textit{view-invariant} FER with the main goal of effective recognition of expressions from any viewing angle. Our method uses a new compound loss function that consists of three specific terms: the first term learns view-invariant representations for facial expressions; the second term learns supervised expression class information, and the third term forces the representations to become compact by learning and de-correlating subtle differences between expressions. We test our proposed framework on two multi-view datasets, KDEF and DDCF, where we achieve state-of-the-art performances when a single-view image is used at inference. Our method shows robustness towards challenging views as the performance does not deviate much from the frontal view. The model also exhibits considerably less reliance on the number of output labels used for training. The proposed method also performs robustly for different encoders. Detailed ablation studies show the impact of each of the components of the proposed method. For future work, the proposed method can be used for other multi-view recognition tasks, for example, multi-view face recognition or multi-view object recognition. Future work can also explore eliminating or reducing the need for knowing the angles during training by using an object or human pose estimator model to estimate the angles, based on which the views could be grouped using the view loss.

\begin{acks}
We would like to thank BMO Bank of Montreal and Mitacs for funding this research. 
\end{acks}
 
\bibliographystyle{ACM-Reference-Format}
\bibliography{references}


\begin{thebibliography}{84}


\ifx \showCODEN    \undefined \def \showCODEN     #1{\unskip}     \fi
\ifx \showDOI      \undefined \def \showDOI       #1{#1}\fi
\ifx \showISBNx    \undefined \def \showISBNx     #1{\unskip}     \fi
\ifx \showISBNxiii \undefined \def \showISBNxiii  #1{\unskip}     \fi
\ifx \showISSN     \undefined \def \showISSN      #1{\unskip}     \fi
\ifx \showLCCN     \undefined \def \showLCCN      #1{\unskip}     \fi
\ifx \shownote     \undefined \def \shownote      #1{#1}          \fi
\ifx \showarticletitle \undefined \def \showarticletitle #1{#1}   \fi
\ifx \showURL      \undefined \def \showURL       {\relax}        \fi
\providecommand\bibfield[2]{#2}
\providecommand\bibinfo[2]{#2}
\providecommand\natexlab[1]{#1}
\providecommand\showeprint[2][]{arXiv:#2}

\bibitem[Albu et~al\mbox{.}(2015)]%
        {albu2015neural}
\bibfield{author}{\bibinfo{person}{Felix Albu}, \bibinfo{person}{Daniela Hagiescu}, \bibinfo{person}{Liviu Vladutu}, {and} \bibinfo{person}{Mihaela-Alexandra Puica}.} \bibinfo{year}{2015}\natexlab{}.
\newblock \showarticletitle{Neural network approaches for children's emotion recognition in intelligent learning applications}. In \bibinfo{booktitle}{\emph{7th Annu Int Conf Educ New Learn Technol Barcelona}}.
\newblock


\bibitem[Bardes et~al\mbox{.}(2021)]%
        {vicreg}
\bibfield{author}{\bibinfo{person}{Adrien Bardes}, \bibinfo{person}{Jean Ponce}, {and} \bibinfo{person}{Yann LeCun}.} \bibinfo{year}{2021}\natexlab{}.
\newblock \showarticletitle{Vicreg: Variance-invariance-covariance regularization for self-supervised learning}.
\newblock \bibinfo{journal}{\emph{arXiv preprint arXiv:2105.04906}} (\bibinfo{year}{2021}).
\newblock


\bibitem[Chen et~al\mbox{.}(2020)]%
        {simclr}
\bibfield{author}{\bibinfo{person}{Ting Chen}, \bibinfo{person}{Simon Kornblith}, \bibinfo{person}{Mohammad Norouzi}, {and} \bibinfo{person}{Geoffrey Hinton}.} \bibinfo{year}{2020}\natexlab{}.
\newblock \showarticletitle{A simple framework for contrastive learning of visual representations}. In \bibinfo{booktitle}{\emph{International conference on machine learning}}. \bibinfo{pages}{1597--1607}.
\newblock


\bibitem[Cho et~al\mbox{.}(2019)]%
        {cho2019instant}
\bibfield{author}{\bibinfo{person}{Youngjun Cho}, \bibinfo{person}{Simon~J Julier}, {and} \bibinfo{person}{Nadia Bianchi-Berthouze}.} \bibinfo{year}{2019}\natexlab{}.
\newblock \showarticletitle{Instant stress: detection of perceived mental stress through smartphone photoplethysmography and thermal imaging}.
\newblock \bibinfo{journal}{\emph{JMIR Mental Health}} \bibinfo{volume}{6}, \bibinfo{number}{4} (\bibinfo{year}{2019}).
\newblock


\bibitem[Chopra et~al\mbox{.}(2005)]%
        {triplet}
\bibfield{author}{\bibinfo{person}{Sumit Chopra}, \bibinfo{person}{Raia Hadsell}, {and} \bibinfo{person}{Yann LeCun}.} \bibinfo{year}{2005}\natexlab{}.
\newblock \showarticletitle{Learning a similarity metric discriminatively, with application to face verification}. In \bibinfo{booktitle}{\emph{2005 IEEE Computer Society Conference on Computer Vision and Pattern Recognition (CVPR'05)}}, Vol.~\bibinfo{volume}{1}. \bibinfo{pages}{539--546}.
\newblock


\bibitem[Dalrymple et~al\mbox{.}(2013)]%
        {ddcf}
\bibfield{author}{\bibinfo{person}{Kirsten~A Dalrymple}, \bibinfo{person}{Jesse Gomez}, {and} \bibinfo{person}{Brad Duchaine}.} \bibinfo{year}{2013}\natexlab{}.
\newblock \showarticletitle{The Dartmouth Database of Children’s Faces: Acquisition and validation of a new face stimulus set}.
\newblock \bibinfo{journal}{\emph{PloS One}} \bibinfo{volume}{8}, \bibinfo{number}{11} (\bibinfo{year}{2013}), \bibinfo{pages}{e79131}.
\newblock


\bibitem[Eleftheriadis et~al\mbox{.}(2014)]%
        {eleftheriadis2014discriminative}
\bibfield{author}{\bibinfo{person}{Stefanos Eleftheriadis}, \bibinfo{person}{Ognjen Rudovic}, {and} \bibinfo{person}{Maja Pantic}.} \bibinfo{year}{2014}\natexlab{}.
\newblock \showarticletitle{Discriminative shared gaussian processes for multiview and view-invariant facial expression recognition}.
\newblock \bibinfo{journal}{\emph{IEEE transactions on image processing}} \bibinfo{volume}{24}, \bibinfo{number}{1} (\bibinfo{year}{2014}), \bibinfo{pages}{189--204}.
\newblock


\bibitem[Essa and Pentland(1995)]%
        {essa1995facial}
\bibfield{author}{\bibinfo{person}{Irfan~A Essa} {and} \bibinfo{person}{Alex~P Pentland}.} \bibinfo{year}{1995}\natexlab{}.
\newblock \showarticletitle{Facial expression recognition using a dynamic model and motion energy}. In \bibinfo{booktitle}{\emph{Proceedings of IEEE International Conference on Computer Vision}}. IEEE, \bibinfo{pages}{360--367}.
\newblock


\bibitem[Fang et~al\mbox{.}(2023)]%
        {fang2023rethinking}
\bibfield{author}{\bibinfo{person}{Bei Fang}, \bibinfo{person}{Xian Li}, \bibinfo{person}{Guangxin Han}, {and} \bibinfo{person}{Juhou He}.} \bibinfo{year}{2023}\natexlab{}.
\newblock \showarticletitle{Rethinking pseudo-labeling for semi-supervised facial expression recognition with contrastive self-supervised learning}.
\newblock \bibinfo{journal}{\emph{IEEE Access}} (\bibinfo{year}{2023}).
\newblock


\bibitem[Fard and Mahoor(2022)]%
        {fard2022ad}
\bibfield{author}{\bibinfo{person}{Ali~Pourramezan Fard} {and} \bibinfo{person}{Mohammad~H Mahoor}.} \bibinfo{year}{2022}\natexlab{}.
\newblock \showarticletitle{Ad-corre: Adaptive correlation-based loss for facial expression recognition in the wild}.
\newblock \bibinfo{journal}{\emph{IEEE Access}}  \bibinfo{volume}{10} (\bibinfo{year}{2022}), \bibinfo{pages}{26756--26768}.
\newblock


\bibitem[Frosst et~al\mbox{.}(2019)]%
        {frosst2019analyzing}
\bibfield{author}{\bibinfo{person}{Nicholas Frosst}, \bibinfo{person}{Nicolas Papernot}, {and} \bibinfo{person}{Geoffrey Hinton}.} \bibinfo{year}{2019}\natexlab{}.
\newblock \showarticletitle{Analyzing and improving representations with the soft nearest neighbor loss}. In \bibinfo{booktitle}{\emph{International Conference on Machine Learning}}. \bibinfo{pages}{2012--2020}.
\newblock


\bibitem[Hariri et~al\mbox{.}(2017)]%
        {hariri20173d}
\bibfield{author}{\bibinfo{person}{Walid Hariri}, \bibinfo{person}{Hedi Tabia}, \bibinfo{person}{Nadir Farah}, \bibinfo{person}{Abdallah Benouareth}, {and} \bibinfo{person}{David Declercq}.} \bibinfo{year}{2017}\natexlab{}.
\newblock \showarticletitle{3D facial expression recognition using kernel methods on Riemannian manifold}.
\newblock \bibinfo{journal}{\emph{Engineering Applications of Artificial Intelligence}}  \bibinfo{volume}{64} (\bibinfo{year}{2017}), \bibinfo{pages}{25--32}.
\newblock


\bibitem[Hasani et~al\mbox{.}(2020)]%
        {hasani2020breg}
\bibfield{author}{\bibinfo{person}{Behzad Hasani}, \bibinfo{person}{Pooran~Singh Negi}, {and} \bibinfo{person}{Mohammad Mahoor}.} \bibinfo{year}{2020}\natexlab{}.
\newblock \showarticletitle{BReG-NeXt: Facial affect computing using adaptive residual networks with bounded gradient}.
\newblock \bibinfo{journal}{\emph{IEEE Transactions on Affective Computing}} (\bibinfo{year}{2020}).
\newblock


\bibitem[He et~al\mbox{.}(2020)]%
        {he2020momentum}
\bibfield{author}{\bibinfo{person}{Kaiming He}, \bibinfo{person}{Haoqi Fan}, \bibinfo{person}{Yuxin Wu}, \bibinfo{person}{Saining Xie}, {and} \bibinfo{person}{Ross Girshick}.} \bibinfo{year}{2020}\natexlab{}.
\newblock \showarticletitle{Momentum contrast for unsupervised visual representation learning}. In \bibinfo{booktitle}{\emph{Proceedings of the IEEE/CVF Conference on Computer Vision and Pattern Recognition}}. \bibinfo{pages}{9729--9738}.
\newblock


\bibitem[He et~al\mbox{.}(2016)]%
        {resnet}
\bibfield{author}{\bibinfo{person}{Kaiming He}, \bibinfo{person}{Xiangyu Zhang}, \bibinfo{person}{Shaoqing Ren}, {and} \bibinfo{person}{Jian Sun}.} \bibinfo{year}{2016}\natexlab{}.
\newblock \showarticletitle{Deep residual learning for image recognition}. In \bibinfo{booktitle}{\emph{Proceedings of the IEEE conference on computer vision and pattern recognition}}. \bibinfo{pages}{770--778}.
\newblock


\bibitem[Henaff(2020)]%
        {henaff2020data}
\bibfield{author}{\bibinfo{person}{Olivier Henaff}.} \bibinfo{year}{2020}\natexlab{}.
\newblock \showarticletitle{Data-efficient image recognition with contrastive predictive coding}. In \bibinfo{booktitle}{\emph{International Conference on Machine Learning}}. \bibinfo{pages}{4182--4192}.
\newblock


\bibitem[Hjelm et~al\mbox{.}(2018)]%
        {hjelm2018learning}
\bibfield{author}{\bibinfo{person}{R~Devon Hjelm}, \bibinfo{person}{Alex Fedorov}, \bibinfo{person}{Samuel Lavoie-Marchildon}, \bibinfo{person}{Karan Grewal}, \bibinfo{person}{Phil Bachman}, \bibinfo{person}{Adam Trischler}, {and} \bibinfo{person}{Yoshua Bengio}.} \bibinfo{year}{2018}\natexlab{}.
\newblock \showarticletitle{Learning deep representations by mutual information estimation and maximization}.
\newblock \bibinfo{journal}{\emph{arXiv preprint arXiv:1808.06670}} (\bibinfo{year}{2018}).
\newblock


\bibitem[Hu et~al\mbox{.}(2008)]%
        {hu2008multi}
\bibfield{author}{\bibinfo{person}{Yuxiao Hu}, \bibinfo{person}{Zhihong Zeng}, \bibinfo{person}{Lijun Yin}, \bibinfo{person}{Xiaozhou Wei}, \bibinfo{person}{Xi Zhou}, {and} \bibinfo{person}{Thomas~S Huang}.} \bibinfo{year}{2008}\natexlab{}.
\newblock \showarticletitle{Multi-view facial expression recognition}. In \bibinfo{booktitle}{\emph{2008 8th IEEE International Conference on Automatic Face \& Gesture Recognition}}. IEEE, \bibinfo{pages}{1--6}.
\newblock


\bibitem[Huang et~al\mbox{.}(2017)]%
        {huang2017densenet}
\bibfield{author}{\bibinfo{person}{Gao Huang}, \bibinfo{person}{Zhuang Liu}, \bibinfo{person}{Laurens Van Der~Maaten}, {and} \bibinfo{person}{Kilian~Q Weinberger}.} \bibinfo{year}{2017}\natexlab{}.
\newblock \showarticletitle{Densely connected convolutional networks}. In \bibinfo{booktitle}{\emph{Proceedings of the IEEE conference on computer vision and pattern recognition}}. \bibinfo{pages}{4700--4708}.
\newblock


\bibitem[Huynh et~al\mbox{.}(2016)]%
        {huynh2016convolutional}
\bibfield{author}{\bibinfo{person}{Xuan-Phung Huynh}, \bibinfo{person}{Tien-Duc Tran}, {and} \bibinfo{person}{Yong-Guk Kim}.} \bibinfo{year}{2016}\natexlab{}.
\newblock \showarticletitle{Convolutional neural network models for facial expression recognition using bu-3dfe database}. In \bibinfo{booktitle}{\emph{Information Science and Applications}}. Springer, \bibinfo{pages}{441--450}.
\newblock


\bibitem[Jiang and Deng(2021)]%
        {jiang2021boosting}
\bibfield{author}{\bibinfo{person}{Jing Jiang} {and} \bibinfo{person}{Weihong Deng}.} \bibinfo{year}{2021}\natexlab{}.
\newblock \showarticletitle{Boosting Facial Expression Recognition by A Semi-Supervised Progressive Teacher}.
\newblock \bibinfo{journal}{\emph{IEEE Transactions on Affective Computing}} (\bibinfo{year}{2021}).
\newblock


\bibitem[Jie and Yongsheng(2020)]%
        {jie2020multi}
\bibfield{author}{\bibinfo{person}{Shao Jie} {and} \bibinfo{person}{Qian Yongsheng}.} \bibinfo{year}{2020}\natexlab{}.
\newblock \showarticletitle{Multi-view facial expression recognition with multi-view facial expression light weight network}.
\newblock \bibinfo{journal}{\emph{Pattern Recognition and Image Analysis}}  \bibinfo{volume}{30} (\bibinfo{year}{2020}), \bibinfo{pages}{805--814}.
\newblock


\bibitem[Khan et~al\mbox{.}(2019)]%
        {khan2019saliency}
\bibfield{author}{\bibinfo{person}{Rizwan~Ahmed Khan}, \bibinfo{person}{Alexandre Meyer}, \bibinfo{person}{Hubert Konik}, {and} \bibinfo{person}{Saida Bouakaz}.} \bibinfo{year}{2019}\natexlab{}.
\newblock \showarticletitle{Saliency-based framework for facial expression recognition}.
\newblock \bibinfo{journal}{\emph{Frontiers of Computer Science}} \bibinfo{volume}{13}, \bibinfo{number}{1} (\bibinfo{year}{2019}), \bibinfo{pages}{183--198}.
\newblock


\bibitem[Khosla et~al\mbox{.}(2020)]%
        {supcon}
\bibfield{author}{\bibinfo{person}{Prannay Khosla}, \bibinfo{person}{Piotr Teterwak}, \bibinfo{person}{Chen Wang}, \bibinfo{person}{Aaron Sarna}, \bibinfo{person}{Yonglong Tian}, \bibinfo{person}{Phillip Isola}, \bibinfo{person}{Aaron Maschinot}, \bibinfo{person}{Ce Liu}, {and} \bibinfo{person}{Dilip Krishnan}.} \bibinfo{year}{2020}\natexlab{}.
\newblock \showarticletitle{Supervised Contrastive Learning}.
\newblock \bibinfo{journal}{\emph{Advances in Neural Information Processing Systems}}  \bibinfo{volume}{33} (\bibinfo{year}{2020}).
\newblock


\bibitem[Kolahdouzi et~al\mbox{.}(2021)]%
        {kolahdouzi2021face}
\bibfield{author}{\bibinfo{person}{Mojtaba Kolahdouzi}, \bibinfo{person}{Alireza Sepas-Moghaddam}, {and} \bibinfo{person}{Ali Etemad}.} \bibinfo{year}{2021}\natexlab{}.
\newblock \showarticletitle{Face Trees for Expression Recognition}. In \bibinfo{booktitle}{\emph{2021 16th IEEE International Conference on Automatic Face and Gesture Recognition (FG 2021)}}. IEEE, \bibinfo{pages}{1--5}.
\newblock


\bibitem[Krizhevsky et~al\mbox{.}(2012)]%
        {krizhevsky2012imagenet}
\bibfield{author}{\bibinfo{person}{Alex Krizhevsky}, \bibinfo{person}{Ilya Sutskever}, {and} \bibinfo{person}{Geoffrey~E Hinton}.} \bibinfo{year}{2012}\natexlab{}.
\newblock \showarticletitle{Imagenet classification with deep convolutional neural networks}.
\newblock \bibinfo{journal}{\emph{Advances in neural information processing systems}}  \bibinfo{volume}{25} (\bibinfo{year}{2012}).
\newblock


\bibitem[Leng et~al\mbox{.}(2007)]%
        {leng2007experimental}
\bibfield{author}{\bibinfo{person}{H Leng}, \bibinfo{person}{Y Lin}, {and} \bibinfo{person}{LA Zanzi}.} \bibinfo{year}{2007}\natexlab{}.
\newblock \showarticletitle{An experimental study on physiological parameters toward driver emotion recognition}. In \bibinfo{booktitle}{\emph{International Conference on Ergonomics and Health Aspects of Work with Computers}}. \bibinfo{pages}{237--246}.
\newblock


\bibitem[Li et~al\mbox{.}(2022b)]%
        {li2022crs}
\bibfield{author}{\bibinfo{person}{Hangyu Li}, \bibinfo{person}{Nannan Wang}, \bibinfo{person}{Xi Yang}, {and} \bibinfo{person}{Xinbo Gao}.} \bibinfo{year}{2022}\natexlab{b}.
\newblock \showarticletitle{CRS-CONT: A Well-Trained General Encoder for Facial Expression Analysis}.
\newblock \bibinfo{journal}{\emph{IEEE Transactions on Image Processing}} (\bibinfo{year}{2022}).
\newblock


\bibitem[Li et~al\mbox{.}(2022c)]%
        {li2022towards}
\bibfield{author}{\bibinfo{person}{Hangyu Li}, \bibinfo{person}{Nannan Wang}, \bibinfo{person}{Xi Yang}, \bibinfo{person}{Xiaoyu Wang}, {and} \bibinfo{person}{Xinbo Gao}.} \bibinfo{year}{2022}\natexlab{c}.
\newblock \showarticletitle{Towards semi-supervised deep facial expression recognition with an adaptive confidence margin}. In \bibinfo{booktitle}{\emph{IEEE/CVF Conference on Computer Vision and Pattern Recognition}}. \bibinfo{pages}{4166--4175}.
\newblock


\bibitem[Li et~al\mbox{.}(2022a)]%
        {li2022optimal}
\bibfield{author}{\bibinfo{person}{Yinqi Li}, \bibinfo{person}{Hong Chang}, \bibinfo{person}{Bingpeng Ma}, \bibinfo{person}{Shiguang Shan}, {and} \bibinfo{person}{Xilin Chen}.} \bibinfo{year}{2022}\natexlab{a}.
\newblock \showarticletitle{Optimal Positive Generation via Latent Transformation for Contrastive Learning}.
\newblock \bibinfo{journal}{\emph{Advances in Neural Information Processing Systems}}  \bibinfo{volume}{35} (\bibinfo{year}{2022}), \bibinfo{pages}{18327--18342}.
\newblock


\bibitem[Li et~al\mbox{.}(2021)]%
        {li2021self}
\bibfield{author}{\bibinfo{person}{Yingjian Li}, \bibinfo{person}{Yingnan Gao}, \bibinfo{person}{Bingzhi Chen}, \bibinfo{person}{Zheng Zhang}, \bibinfo{person}{Guangming Lu}, {and} \bibinfo{person}{David Zhang}.} \bibinfo{year}{2021}\natexlab{}.
\newblock \showarticletitle{Self-supervised exclusive-inclusive interactive learning for multi-label facial expression recognition in the wild}.
\newblock \bibinfo{journal}{\emph{IEEE Transactions on Circuits and Systems for Video Technology}} (\bibinfo{year}{2021}).
\newblock


\bibitem[Liu et~al\mbox{.}(2023)]%
        {liu2023soft}
\bibfield{author}{\bibinfo{person}{Chaoji Liu}, \bibinfo{person}{Xingqiao Liu}, \bibinfo{person}{Chong Chen}, {and} \bibinfo{person}{Qiankun Wang}.} \bibinfo{year}{2023}\natexlab{}.
\newblock \showarticletitle{Soft thresholding squeeze-and-excitation network for pose-invariant facial expression recognition}.
\newblock \bibinfo{journal}{\emph{The Visual Computer}} \bibinfo{volume}{39}, \bibinfo{number}{7} (\bibinfo{year}{2023}), \bibinfo{pages}{2637--2652}.
\newblock


\bibitem[Liu et~al\mbox{.}(2019)]%
        {PhaNet}
\bibfield{author}{\bibinfo{person}{Yuanyuan Liu}, \bibinfo{person}{Jiyao Peng}, \bibinfo{person}{Jiabei Zeng}, {and} \bibinfo{person}{Shiguang Shan}.} \bibinfo{year}{2019}\natexlab{}.
\newblock \showarticletitle{Pose-adaptive Hierarchical Attention Network for Facial Expression Recognition}.
\newblock \bibinfo{journal}{\emph{arXiv preprint arXiv:1905.10059}} (\bibinfo{year}{2019}).
\newblock


\bibitem[Liu et~al\mbox{.}(2018)]%
        {MPCNN}
\bibfield{author}{\bibinfo{person}{Yuanyuan Liu}, \bibinfo{person}{Jiabei Zeng}, \bibinfo{person}{Shiguang Shan}, {and} \bibinfo{person}{Zhuo Zheng}.} \bibinfo{year}{2018}\natexlab{}.
\newblock \showarticletitle{Multi-channel pose-aware convolution neural networks for multi-view facial expression recognition}. In \bibinfo{booktitle}{\emph{13th IEEE International Conference on Automatic Face \& Gesture Recognition}}. \bibinfo{pages}{458--465}.
\newblock


\bibitem[Lundqvist et~al\mbox{.}(1998)]%
        {kdef}
\bibfield{author}{\bibinfo{person}{Daniel Lundqvist}, \bibinfo{person}{Anders Flykt}, {and} \bibinfo{person}{Arne {\"O}hman}.} \bibinfo{year}{1998}\natexlab{}.
\newblock \showarticletitle{The Karolinska Directed Emotional Faces (KDEF)}.
\newblock \bibinfo{journal}{\emph{CD ROM from Department of Clinical Neuroscience, Psychology Section, Karolinska Institutet}} \bibinfo{volume}{91}, \bibinfo{number}{630} (\bibinfo{year}{1998}), \bibinfo{pages}{2--2}.
\newblock


\bibitem[Mahesh et~al\mbox{.}(2021)]%
        {RB-FNN}
\bibfield{author}{\bibinfo{person}{Vijayalakshmi~GV Mahesh}, \bibinfo{person}{Chengji Chen}, \bibinfo{person}{Vijayarajan Rajangam}, \bibinfo{person}{Alex Noel~Joseph Raj}, {and} \bibinfo{person}{Palani~Thanaraj Krishnan}.} \bibinfo{year}{2021}\natexlab{}.
\newblock \showarticletitle{Shape and Texture Aware Facial Expression Recognition Using Spatial Pyramid Zernike Moments and Law’s Textures Feature Set}.
\newblock \bibinfo{journal}{\emph{IEEE Access}}  \bibinfo{volume}{9} (\bibinfo{year}{2021}), \bibinfo{pages}{52509--52522}.
\newblock


\bibitem[Misra and Maaten(2020)]%
        {misra2020self}
\bibfield{author}{\bibinfo{person}{Ishan Misra} {and} \bibinfo{person}{Laurens van~der Maaten}.} \bibinfo{year}{2020}\natexlab{}.
\newblock \showarticletitle{Self-supervised learning of pretext-invariant representations}. In \bibinfo{booktitle}{\emph{Proceedings of the IEEE/CVF Conference on Computer Vision and Pattern Recognition}}. \bibinfo{pages}{6707--6717}.
\newblock


\bibitem[Miyai et~al\mbox{.}(2023)]%
        {miyai2023rethinking}
\bibfield{author}{\bibinfo{person}{Atsuyuki Miyai}, \bibinfo{person}{Qing Yu}, \bibinfo{person}{Daiki Ikami}, \bibinfo{person}{Go Irie}, {and} \bibinfo{person}{Kiyoharu Aizawa}.} \bibinfo{year}{2023}\natexlab{}.
\newblock \showarticletitle{Rethinking Rotation in Self-Supervised Contrastive Learning: Adaptive Positive or Negative Data Augmentation}. In \bibinfo{booktitle}{\emph{IEEE/CVF Winter Conference on Applications of Computer Vision}}. \bibinfo{pages}{2809--2818}.
\newblock


\bibitem[Moore and Bowden(2011)]%
        {SVM}
\bibfield{author}{\bibinfo{person}{Stephen Moore} {and} \bibinfo{person}{Richard Bowden}.} \bibinfo{year}{2011}\natexlab{}.
\newblock \showarticletitle{Local binary patterns for multi-view facial expression recognition}.
\newblock \bibinfo{journal}{\emph{Computer Vision and Image Understanding}} \bibinfo{volume}{115}, \bibinfo{number}{4} (\bibinfo{year}{2011}), \bibinfo{pages}{541--558}.
\newblock


\bibitem[Pourmirzaei et~al\mbox{.}(2021)]%
        {pourmirzaei2021using}
\bibfield{author}{\bibinfo{person}{Mahdi Pourmirzaei}, \bibinfo{person}{Farzaneh Esmaili}, {and} \bibinfo{person}{Gholam~Ali Montazer}.} \bibinfo{year}{2021}\natexlab{}.
\newblock \showarticletitle{Using Self-Supervised Co-Training to Improve Facial Representation}.
\newblock \bibinfo{journal}{\emph{arXiv preprint arXiv:2105.06421}} (\bibinfo{year}{2021}).
\newblock


\bibitem[Psaroudakis and Kollias(2022)]%
        {psaroudakis2022mixaugment}
\bibfield{author}{\bibinfo{person}{Andreas Psaroudakis} {and} \bibinfo{person}{Dimitrios Kollias}.} \bibinfo{year}{2022}\natexlab{}.
\newblock \showarticletitle{Mixaugment \& mixup: Augmentation methods for facial expression recognition}. In \bibinfo{booktitle}{\emph{IEEE/CVF Conference on Computer Vision and Pattern Recognition}}. \bibinfo{pages}{2367--2375}.
\newblock


\bibitem[Rao et~al\mbox{.}(2015)]%
        {SURF}
\bibfield{author}{\bibinfo{person}{Qiyu Rao}, \bibinfo{person}{Xing Qu}, \bibinfo{person}{Qirong Mao}, {and} \bibinfo{person}{Yongzhao Zhan}.} \bibinfo{year}{2015}\natexlab{}.
\newblock \showarticletitle{Multi-pose facial expression recognition based on SURF boosting}. In \bibinfo{booktitle}{\emph{IEEE International Conference on Affective Computing and Intelligent Interaction}}. \bibinfo{pages}{630--635}.
\newblock


\bibitem[Roy and Etemad(2021a)]%
        {CL_MEx}
\bibfield{author}{\bibinfo{person}{Shuvendu Roy} {and} \bibinfo{person}{Ali Etemad}.} \bibinfo{year}{2021}\natexlab{a}.
\newblock \showarticletitle{Self-supervised contrastive learning of multi-view facial expressions}. In \bibinfo{booktitle}{\emph{Proceedings of the 2021 International Conference on Multimodal Interaction}}. \bibinfo{pages}{253--257}.
\newblock


\bibitem[Roy and Etemad(2021b)]%
        {ST-CLR}
\bibfield{author}{\bibinfo{person}{Shuvendu Roy} {and} \bibinfo{person}{Ali Etemad}.} \bibinfo{year}{2021}\natexlab{b}.
\newblock \showarticletitle{Spatiotemporal Contrastive Learning of Facial Expressions in Videos}.
\newblock  (\bibinfo{year}{2021}), \bibinfo{pages}{1--8}.
\newblock


\bibitem[Roy and Etemad(2023a)]%
        {roy2023active}
\bibfield{author}{\bibinfo{person}{Shuvendu Roy} {and} \bibinfo{person}{Ali Etemad}.} \bibinfo{year}{2023}\natexlab{a}.
\newblock \showarticletitle{Active Learning with Contrastive Pre-training for Facial Expression Recognition}. In \bibinfo{booktitle}{\emph{International Conference on Affective Computing and Intelligent Interaction}}.
\newblock


\bibitem[Roy and Etemad(2023b)]%
        {concur}
\bibfield{author}{\bibinfo{person}{Shuvendu Roy} {and} \bibinfo{person}{Ali Etemad}.} \bibinfo{year}{2023}\natexlab{b}.
\newblock \showarticletitle{Temporal Contrastive Learning with Curriculum}. In \bibinfo{booktitle}{\emph{IEEE International Conference on Acoustics, Speech and Signal Processing}}. \bibinfo{pages}{1--5}.
\newblock


\bibitem[Salakhutdinov and Hinton(2007)]%
        {salakhutdinov2007learning}
\bibfield{author}{\bibinfo{person}{Ruslan Salakhutdinov} {and} \bibinfo{person}{Geoff Hinton}.} \bibinfo{year}{2007}\natexlab{}.
\newblock \showarticletitle{Learning a nonlinear embedding by preserving class neighbourhood structure}. In \bibinfo{booktitle}{\emph{Artificial Intelligence and Statistics}}. \bibinfo{pages}{412--419}.
\newblock


\bibitem[Samal and Iyengar(1992)]%
        {samal1992automatic}
\bibfield{author}{\bibinfo{person}{Ashok Samal} {and} \bibinfo{person}{Prasana~A Iyengar}.} \bibinfo{year}{1992}\natexlab{}.
\newblock \showarticletitle{Automatic recognition and analysis of human faces and facial expressions: A survey}.
\newblock \bibinfo{journal}{\emph{Pattern recognition}} \bibinfo{volume}{25}, \bibinfo{number}{1} (\bibinfo{year}{1992}), \bibinfo{pages}{65--77}.
\newblock


\bibitem[Sanchez-Cortes et~al\mbox{.}(2013)]%
        {sanchez2013inferring}
\bibfield{author}{\bibinfo{person}{Dairazalia Sanchez-Cortes}, \bibinfo{person}{Joan-Isaac Biel}, \bibinfo{person}{Shiro Kumano}, \bibinfo{person}{Junji Yamato}, \bibinfo{person}{Kazuhiro Otsuka}, {and} \bibinfo{person}{Daniel Gatica-Perez}.} \bibinfo{year}{2013}\natexlab{}.
\newblock \showarticletitle{Inferring mood in ubiquitous conversational video}. In \bibinfo{booktitle}{\emph{12th International Conference on Mobile and Ubiquitous Multimedia}}. \bibinfo{pages}{1--9}.
\newblock


\bibitem[Santra and Mukherjee(2016)]%
        {santra2016local}
\bibfield{author}{\bibinfo{person}{Bikash Santra} {and} \bibinfo{person}{Dipti~Prasad Mukherjee}.} \bibinfo{year}{2016}\natexlab{}.
\newblock \showarticletitle{Local dominant binary patterns for recognition of multi-view facial expressions}. In \bibinfo{booktitle}{\emph{Proceedings of the Tenth Indian Conference on Computer Vision, Graphics and Image Processing}}. \bibinfo{pages}{1--8}.
\newblock


\bibitem[Schoneveld et~al\mbox{.}(2021)]%
        {schoneveld2021leveraging}
\bibfield{author}{\bibinfo{person}{Liam Schoneveld}, \bibinfo{person}{Alice Othmani}, {and} \bibinfo{person}{Hazem Abdelkawy}.} \bibinfo{year}{2021}\natexlab{}.
\newblock \showarticletitle{Leveraging recent advances in deep learning for audio-Visual emotion recognition}.
\newblock \bibinfo{journal}{\emph{Pattern Recognition Letters}} (\bibinfo{year}{2021}).
\newblock


\bibitem[Sepas-Moghaddam et~al\mbox{.}(2019)]%
        {sepas2019deep}
\bibfield{author}{\bibinfo{person}{Alireza Sepas-Moghaddam}, \bibinfo{person}{Ali Etemad}, \bibinfo{person}{Paulo~Lobato Correia}, {and} \bibinfo{person}{Fernando Pereira}.} \bibinfo{year}{2019}\natexlab{}.
\newblock \showarticletitle{A deep framework for facial emotion recognition using light field images}. In \bibinfo{booktitle}{\emph{2019 8th International Conference on Affective Computing and Intelligent Interaction (ACII)}}. IEEE, \bibinfo{pages}{1--7}.
\newblock


\bibitem[Sepas-Moghaddam et~al\mbox{.}(2020a)]%
        {sepas2020facial}
\bibfield{author}{\bibinfo{person}{Alireza Sepas-Moghaddam}, \bibinfo{person}{Ali Etemad}, \bibinfo{person}{Fernando Pereira}, {and} \bibinfo{person}{Paulo~Lobato Correia}.} \bibinfo{year}{2020}\natexlab{a}.
\newblock \showarticletitle{Facial emotion recognition using light field images with deep attention-based bidirectional LSTM}. In \bibinfo{booktitle}{\emph{IEEE International Conference on Acoustics, Speech and Signal Processing (ICASSP)}}. \bibinfo{pages}{3367--3371}.
\newblock


\bibitem[Sepas-Moghaddam et~al\mbox{.}(2020b)]%
        {sepas2020long}
\bibfield{author}{\bibinfo{person}{Alireza Sepas-Moghaddam}, \bibinfo{person}{Ali Etemad}, \bibinfo{person}{Fernando Pereira}, {and} \bibinfo{person}{Paulo~Lobato Correia}.} \bibinfo{year}{2020}\natexlab{b}.
\newblock \showarticletitle{Long short-term memory with gate and state level fusion for light field-based face recognition}.
\newblock \bibinfo{journal}{\emph{IEEE Transactions on Information Forensics and Security}}  \bibinfo{volume}{16} (\bibinfo{year}{2020}), \bibinfo{pages}{1365--1379}.
\newblock


\bibitem[Sepas-Moghaddam et~al\mbox{.}(2021a)]%
        {sepas2021capsfield}
\bibfield{author}{\bibinfo{person}{Alireza Sepas-Moghaddam}, \bibinfo{person}{Ali Etemad}, \bibinfo{person}{Fernando Pereira}, {and} \bibinfo{person}{Paulo~Lobato Correia}.} \bibinfo{year}{2021}\natexlab{a}.
\newblock \showarticletitle{Capsfield: Light field-based face and expression recognition in the wild using capsule routing}.
\newblock \bibinfo{journal}{\emph{IEEE Transactions on Image Processing}}  \bibinfo{volume}{30} (\bibinfo{year}{2021}), \bibinfo{pages}{2627--2642}.
\newblock


\bibitem[Sepas-Moghaddam et~al\mbox{.}(2021b)]%
        {sepas2021multi}
\bibfield{author}{\bibinfo{person}{Alireza Sepas-Moghaddam}, \bibinfo{person}{Fernando Pereira}, \bibinfo{person}{Paulo~Lobato Correia}, {and} \bibinfo{person}{Ali Etemad}.} \bibinfo{year}{2021}\natexlab{b}.
\newblock \showarticletitle{Multi-Perspective LSTM for Joint Visual Representation Learning}. In \bibinfo{booktitle}{\emph{Proceedings of the IEEE/CVF Conference on Computer Vision and Pattern Recognition}}. \bibinfo{pages}{16540--16548}.
\newblock


\bibitem[Sermanet et~al\mbox{.}(2018)]%
        {sermanet2018time}
\bibfield{author}{\bibinfo{person}{Pierre Sermanet}, \bibinfo{person}{Corey Lynch}, \bibinfo{person}{Yevgen Chebotar}, \bibinfo{person}{Jasmine Hsu}, \bibinfo{person}{Eric Jang}, \bibinfo{person}{Stefan Schaal}, \bibinfo{person}{Sergey Levine}, {and} \bibinfo{person}{Google Brain}.} \bibinfo{year}{2018}\natexlab{}.
\newblock \showarticletitle{Time-contrastive networks: Self-supervised learning from video}. In \bibinfo{booktitle}{\emph{IEEE international conference on robotics and automation (ICRA)}}. \bibinfo{pages}{1134--1141}.
\newblock


\bibitem[Shan et~al\mbox{.}(2005)]%
        {shan2005robust}
\bibfield{author}{\bibinfo{person}{Caifeng Shan}, \bibinfo{person}{Shaogang Gong}, {and} \bibinfo{person}{Peter~W McOwan}.} \bibinfo{year}{2005}\natexlab{}.
\newblock \showarticletitle{Robust facial expression recognition using local binary patterns}. In \bibinfo{booktitle}{\emph{IEEE International Conference on Image Processing 2005}}, Vol.~\bibinfo{volume}{2}. IEEE, \bibinfo{pages}{II--370}.
\newblock


\bibitem[Shan et~al\mbox{.}(2009)]%
        {shan2009facial}
\bibfield{author}{\bibinfo{person}{Caifeng Shan}, \bibinfo{person}{Shaogang Gong}, {and} \bibinfo{person}{Peter~W McOwan}.} \bibinfo{year}{2009}\natexlab{}.
\newblock \showarticletitle{Facial expression recognition based on local binary patterns: A comprehensive study}.
\newblock \bibinfo{journal}{\emph{Image and vision Computing}} \bibinfo{volume}{27}, \bibinfo{number}{6} (\bibinfo{year}{2009}), \bibinfo{pages}{803--816}.
\newblock


\bibitem[Simonyan and Zisserman(2014)]%
        {simonyan2014very}
\bibfield{author}{\bibinfo{person}{Karen Simonyan} {and} \bibinfo{person}{Andrew Zisserman}.} \bibinfo{year}{2014}\natexlab{}.
\newblock \showarticletitle{Very deep convolutional networks for large-scale image recognition}.
\newblock \bibinfo{journal}{\emph{arXiv preprint arXiv:1409.1556}} (\bibinfo{year}{2014}).
\newblock


\bibitem[Siqueira et~al\mbox{.}(2020)]%
        {siqueira2020efficient}
\bibfield{author}{\bibinfo{person}{Henrique Siqueira}, \bibinfo{person}{Sven Magg}, {and} \bibinfo{person}{Stefan Wermter}.} \bibinfo{year}{2020}\natexlab{}.
\newblock \showarticletitle{Efficient facial feature learning with wide ensemble-based convolutional neural networks}. In \bibinfo{booktitle}{\emph{AAAI conference on artificial intelligence}}, Vol.~\bibinfo{volume}{34}. \bibinfo{pages}{5800--5809}.
\newblock


\bibitem[Szegedy et~al\mbox{.}(2015)]%
        {inception}
\bibfield{author}{\bibinfo{person}{Christian Szegedy}, \bibinfo{person}{Wei Liu}, \bibinfo{person}{Yangqing Jia}, \bibinfo{person}{Pierre Sermanet}, \bibinfo{person}{Scott Reed}, \bibinfo{person}{Dragomir Anguelov}, \bibinfo{person}{Dumitru Erhan}, \bibinfo{person}{Vincent Vanhoucke}, {and} \bibinfo{person}{Andrew Rabinovich}.} \bibinfo{year}{2015}\natexlab{}.
\newblock \showarticletitle{Going deeper with convolutions}. In \bibinfo{booktitle}{\emph{Proceedings of the IEEE conference on computer vision and pattern recognition}}. \bibinfo{pages}{1--9}.
\newblock


\bibitem[Taheri et~al\mbox{.}(2011)]%
        {taheri2011towards}
\bibfield{author}{\bibinfo{person}{Sima Taheri}, \bibinfo{person}{Pavan Turaga}, {and} \bibinfo{person}{Rama Chellappa}.} \bibinfo{year}{2011}\natexlab{}.
\newblock \showarticletitle{Towards view-invariant expression analysis using analytic shape manifolds}. In \bibinfo{booktitle}{\emph{2011 IEEE International Conference on Automatic Face \& Gesture Recognition (FG)}}. IEEE, \bibinfo{pages}{306--313}.
\newblock


\bibitem[Tang et~al\mbox{.}(2020)]%
        {tang2020facial}
\bibfield{author}{\bibinfo{person}{Yan Tang}, \bibinfo{person}{Xingming Zhang}, \bibinfo{person}{Xiping Hu}, \bibinfo{person}{Siqi Wang}, {and} \bibinfo{person}{Haoxiang Wang}.} \bibinfo{year}{2020}\natexlab{}.
\newblock \showarticletitle{Facial expression recognition using frequency neural network}.
\newblock \bibinfo{journal}{\emph{IEEE Transactions on Image Processing}}  \bibinfo{volume}{30} (\bibinfo{year}{2020}), \bibinfo{pages}{444--457}.
\newblock


\bibitem[Thrasher et~al\mbox{.}(2011)]%
        {thrasher2011mood}
\bibfield{author}{\bibinfo{person}{Michelle Thrasher}, \bibinfo{person}{Marjolein~D Van~der Zwaag}, \bibinfo{person}{Nadia Bianchi-Berthouze}, {and} \bibinfo{person}{Joyce~HDM Westerink}.} \bibinfo{year}{2011}\natexlab{}.
\newblock \showarticletitle{Mood recognition based on upper body posture and movement features}. In \bibinfo{booktitle}{\emph{International Conference on Affective Computing and Intelligent Interaction}}. \bibinfo{pages}{377--386}.
\newblock


\bibitem[Tian et~al\mbox{.}(2020)]%
        {tian2020contrastive}
\bibfield{author}{\bibinfo{person}{Yonglong Tian}, \bibinfo{person}{Dilip Krishnan}, {and} \bibinfo{person}{Phillip Isola}.} \bibinfo{year}{2020}\natexlab{}.
\newblock \showarticletitle{Contrastive multiview coding}. In \bibinfo{booktitle}{\emph{Proceedings of the European Conference on Computer Vision}}. \bibinfo{pages}{776--794}.
\newblock


\bibitem[Tokuno et~al\mbox{.}(2011)]%
        {tokuno2011usage}
\bibfield{author}{\bibinfo{person}{Shinichi Tokuno}, \bibinfo{person}{Gentaro Tsumatori}, \bibinfo{person}{Satoshi Shono}, \bibinfo{person}{Eriko Takei}, \bibinfo{person}{Taisuke Yamamoto}, \bibinfo{person}{Go Suzuki}, \bibinfo{person}{Shunnji Mituyoshi}, {and} \bibinfo{person}{Makoto Shimura}.} \bibinfo{year}{2011}\natexlab{}.
\newblock \showarticletitle{Usage of emotion recognition in military health care}. In \bibinfo{booktitle}{\emph{Defense Science Research Conference and Expo}}. \bibinfo{pages}{1--5}.
\newblock


\bibitem[Venkatesh et~al\mbox{.}(2012)]%
        {venkatesh2012simultaneous}
\bibfield{author}{\bibinfo{person}{Yedatore~V Venkatesh}, \bibinfo{person}{Ashraf~A Kassim}, \bibinfo{person}{Jun Yuan}, {and} \bibinfo{person}{Tan~Dat Nguyen}.} \bibinfo{year}{2012}\natexlab{}.
\newblock \showarticletitle{On the simultaneous recognition of identity and expression from BU-3DFE datasets}.
\newblock \bibinfo{journal}{\emph{Pattern Recognition Letters}} \bibinfo{volume}{33}, \bibinfo{number}{13} (\bibinfo{year}{2012}), \bibinfo{pages}{1785--1793}.
\newblock


\bibitem[Vo et~al\mbox{.}(2019)]%
        {vo20193d}
\bibfield{author}{\bibinfo{person}{Quang~Nhat Vo}, \bibinfo{person}{Khanh Tran}, {and} \bibinfo{person}{Guoying Zhao}.} \bibinfo{year}{2019}\natexlab{}.
\newblock \showarticletitle{3d facial expression recognition based on multi-view and prior knowledge fusion}. In \bibinfo{booktitle}{\emph{2019 IEEE 21st International Workshop on Multimedia Signal Processing (MMSP)}}. IEEE, \bibinfo{pages}{1--6}.
\newblock


\bibitem[Vo et~al\mbox{.}(2020)]%
        {vo2020pyramid}
\bibfield{author}{\bibinfo{person}{Thanh-Hung Vo}, \bibinfo{person}{Guee-Sang Lee}, \bibinfo{person}{Hyung-Jeong Yang}, {and} \bibinfo{person}{Soo-Hyung Kim}.} \bibinfo{year}{2020}\natexlab{}.
\newblock \showarticletitle{Pyramid with super resolution for In-the-Wild facial expression recognition}.
\newblock \bibinfo{journal}{\emph{IEEE Access}}  \bibinfo{volume}{8} (\bibinfo{year}{2020}), \bibinfo{pages}{131988--132001}.
\newblock


\bibitem[Wang et~al\mbox{.}(2020)]%
        {wang2020region}
\bibfield{author}{\bibinfo{person}{Kai Wang}, \bibinfo{person}{Xiaojiang Peng}, \bibinfo{person}{Jianfei Yang}, \bibinfo{person}{Debin Meng}, {and} \bibinfo{person}{Yu Qiao}.} \bibinfo{year}{2020}\natexlab{}.
\newblock \showarticletitle{Region attention networks for pose and occlusion robust facial expression recognition}.
\newblock \bibinfo{journal}{\emph{IEEE Transactions on Image Processing}}  \bibinfo{volume}{29} (\bibinfo{year}{2020}), \bibinfo{pages}{4057--4069}.
\newblock


\bibitem[Wang et~al\mbox{.}(2021)]%
        {wang2021multi}
\bibfield{author}{\bibinfo{person}{Lingfeng Wang}, \bibinfo{person}{Shisen Wang}, \bibinfo{person}{Jin Qi}, {and} \bibinfo{person}{Kenji Suzuki}.} \bibinfo{year}{2021}\natexlab{}.
\newblock \showarticletitle{A multi-task mean teacher for semi-supervised facial affective behavior analysis}. In \bibinfo{booktitle}{\emph{Proceedings of the IEEE/CVF International Conference on Computer Vision}}. \bibinfo{pages}{3603--3608}.
\newblock


\bibitem[Wang et~al\mbox{.}(2016)]%
        {wang2016facial}
\bibfield{author}{\bibinfo{person}{Yiming Wang}, \bibinfo{person}{Hui Yu}, \bibinfo{person}{Junyu Dong}, \bibinfo{person}{Brett Stevens}, {and} \bibinfo{person}{Honghai Liu}.} \bibinfo{year}{2016}\natexlab{}.
\newblock \showarticletitle{Facial expression-aware face frontalization}. In \bibinfo{booktitle}{\emph{Asian Conference on Computer Vision}}. Springer, \bibinfo{pages}{375--388}.
\newblock


\bibitem[Wu et~al\mbox{.}(2018a)]%
        {wu2018improving}
\bibfield{author}{\bibinfo{person}{Zhirong Wu}, \bibinfo{person}{Alexei~A Efros}, {and} \bibinfo{person}{Stella~X Yu}.} \bibinfo{year}{2018}\natexlab{a}.
\newblock \showarticletitle{Improving generalization via scalable neighborhood component analysis}. In \bibinfo{booktitle}{\emph{Proceedings of the European Conference on Computer Vision}}. \bibinfo{pages}{685--701}.
\newblock


\bibitem[Wu et~al\mbox{.}(2018b)]%
        {wu2018unsupervised}
\bibfield{author}{\bibinfo{person}{Zhirong Wu}, \bibinfo{person}{Yuanjun Xiong}, \bibinfo{person}{Stella~X Yu}, {and} \bibinfo{person}{Dahua Lin}.} \bibinfo{year}{2018}\natexlab{b}.
\newblock \showarticletitle{Unsupervised feature learning via non-parametric instance discrimination}. In \bibinfo{booktitle}{\emph{Proceedings of the IEEE conference on computer vision and pattern recognition}}. \bibinfo{pages}{3733--3742}.
\newblock


\bibitem[Xia and Wang(2021)]%
        {xia2021micro}
\bibfield{author}{\bibinfo{person}{Bin Xia} {and} \bibinfo{person}{Shangfei Wang}.} \bibinfo{year}{2021}\natexlab{}.
\newblock \showarticletitle{Micro-Expression Recognition Enhanced by Macro-Expression from Spatial-Temporal Domain.}. In \bibinfo{booktitle}{\emph{International Joint Conference on Artificial Intelligence}}. \bibinfo{pages}{1186--1193}.
\newblock


\bibitem[Xue et~al\mbox{.}(2022)]%
        {xue2022coarse}
\bibfield{author}{\bibinfo{person}{Fanglei Xue}, \bibinfo{person}{Zichang Tan}, \bibinfo{person}{Yu Zhu}, \bibinfo{person}{Zhongsong Ma}, {and} \bibinfo{person}{Guodong Guo}.} \bibinfo{year}{2022}\natexlab{}.
\newblock \showarticletitle{Coarse-to-fine cascaded networks with smooth predicting for video facial expression recognition}. In \bibinfo{booktitle}{\emph{IEEE/CVF Conference on Computer Vision and Pattern Recognition}}. \bibinfo{pages}{2412--2418}.
\newblock


\bibitem[Yin et~al\mbox{.}(2006)]%
        {bu3d}
\bibfield{author}{\bibinfo{person}{Lijun Yin}, \bibinfo{person}{Xiaozhou Wei}, \bibinfo{person}{Yi Sun}, \bibinfo{person}{Jun Wang}, {and} \bibinfo{person}{Matthew~J Rosato}.} \bibinfo{year}{2006}\natexlab{}.
\newblock \showarticletitle{A 3D facial expression database for facial behavior research}. In \bibinfo{booktitle}{\emph{International Conference on Automatic Face and Gesture Recognition}}. \bibinfo{pages}{211--216}.
\newblock


\bibitem[Zbontar et~al\mbox{.}(2021)]%
        {barlowTwin}
\bibfield{author}{\bibinfo{person}{Jure Zbontar}, \bibinfo{person}{Li Jing}, \bibinfo{person}{Ishan Misra}, \bibinfo{person}{Yann LeCun}, {and} \bibinfo{person}{St{\'e}phane Deny}.} \bibinfo{year}{2021}\natexlab{}.
\newblock \showarticletitle{Barlow twins: Self-supervised learning via redundancy reduction}.
\newblock \bibinfo{journal}{\emph{arXiv preprint arXiv:2103.03230}} (\bibinfo{year}{2021}).
\newblock


\bibitem[Zeng et~al\mbox{.}(2022)]%
        {zeng2022face2exp}
\bibfield{author}{\bibinfo{person}{Dan Zeng}, \bibinfo{person}{Zhiyuan Lin}, \bibinfo{person}{Xiao Yan}, \bibinfo{person}{Yuting Liu}, \bibinfo{person}{Fei Wang}, {and} \bibinfo{person}{Bo Tang}.} \bibinfo{year}{2022}\natexlab{}.
\newblock \showarticletitle{Face2exp: Combating data biases for facial expression recognition}. In \bibinfo{booktitle}{\emph{IEEE/CVF Conference on Computer Vision and Pattern Recognition}}. \bibinfo{pages}{20291--20300}.
\newblock


\bibitem[Zhang et~al\mbox{.}(2018)]%
        {zhang2018spatially}
\bibfield{author}{\bibinfo{person}{Feifei Zhang}, \bibinfo{person}{Qirong Mao}, \bibinfo{person}{Xiangjun Shen}, \bibinfo{person}{Yongzhao Zhan}, {and} \bibinfo{person}{Ming Dong}.} \bibinfo{year}{2018}\natexlab{}.
\newblock \showarticletitle{Spatially coherent feature learning for pose-invariant facial expression recognition}.
\newblock \bibinfo{journal}{\emph{ACM Transactions on Multimedia Computing, Communications, and Applications (TOMM)}} \bibinfo{volume}{14}, \bibinfo{number}{1s} (\bibinfo{year}{2018}), \bibinfo{pages}{1--19}.
\newblock


\bibitem[Zhang et~al\mbox{.}(2020)]%
        {zhang2020geometry}
\bibfield{author}{\bibinfo{person}{Feifei Zhang}, \bibinfo{person}{Tianzhu Zhang}, \bibinfo{person}{Qirong Mao}, {and} \bibinfo{person}{Changsheng Xu}.} \bibinfo{year}{2020}\natexlab{}.
\newblock \showarticletitle{Geometry guided pose-invariant facial expression recognition}.
\newblock \bibinfo{journal}{\emph{IEEE Transactions on Image Processing}}  \bibinfo{volume}{29} (\bibinfo{year}{2020}), \bibinfo{pages}{4445--4460}.
\newblock


\bibitem[Zheng et~al\mbox{.}(2021)]%
        {zheng2021weakly}
\bibfield{author}{\bibinfo{person}{Mingkai Zheng}, \bibinfo{person}{Fei Wang}, \bibinfo{person}{Shan You}, \bibinfo{person}{Chen Qian}, \bibinfo{person}{Changshui Zhang}, \bibinfo{person}{Xiaogang Wang}, {and} \bibinfo{person}{Chang Xu}.} \bibinfo{year}{2021}\natexlab{}.
\newblock \showarticletitle{Weakly supervised contrastive learning}. In \bibinfo{booktitle}{\emph{Proceedings of the IEEE/CVF International Conference on Computer Vision}}. \bibinfo{pages}{10042--10051}.
\newblock


\bibitem[Zhou and Shi(2017)]%
        {TLCNN}
\bibfield{author}{\bibinfo{person}{Yuqian Zhou} {and} \bibinfo{person}{Bertram~E Shi}.} \bibinfo{year}{2017}\natexlab{}.
\newblock \showarticletitle{Action unit selective feature maps in deep networks for facial expression recognition}. In \bibinfo{booktitle}{\emph{IEEE International Joint Conference on Neural Networks}}. \bibinfo{pages}{2031--2038}.
\newblock


\end{thebibliography}

\end{document}